\documentclass{ieeeaccess}
\usepackage{cite}
\usepackage{amsmath,amssymb,amsfonts}
\usepackage{graphicx}
\usepackage{textcomp}
\usepackage{multirow}
\usepackage{bm}
\usepackage{array}
\usepackage{hyperref}
\usepackage[nameinlink,capitalize]{cleveref}
\usepackage[labelformat=simple]{subcaption}

\usepackage{algorithm}
\usepackage{algpseudocode}

\newcommand{\etal}{\textit{et al}.}

\def\BibTeX{{\rm B\kern-.05em{\sc i\kern-.025em b}\kern-.08em
    T\kern-.1667em\lower.7ex\hbox{E}\kern-.125emX}}
\begin{document}
\history{Date of publication xxxx 00, 0000, date of current version xxxx 00, 0000.}
\doi{10.1109/ACCESS.2023.0322000}

\title{Word-level Sign Language Recognition with Multi-stream Neural Networks Focusing on Local Regions and Skeletal Information}
\author{\uppercase{Mizuki Maruyama}\authorrefmark{1}, 
\uppercase{Shrey Singh}\authorrefmark{2},
\uppercase{Katsufumi Inoue}\authorrefmark{3},\IEEEmembership{Member, IEEE},\\
\uppercase{Partha Pratim Roy}\authorrefmark{2},\IEEEmembership{Member, IEEE},
\uppercase{Masakazu Iwamura}\authorrefmark{3},\IEEEmembership{Member, IEEE}, and 
\uppercase{Michifumi Yoshioka}\authorrefmark{3}
}

\address[1]{Graduate School of Engineering, Osaka Prefecture University, 1-1 Gakuencho, Naka, Sakai, Osaka 599-8531 Japan}
\address[2]{Department of Computer Science and Engineering, Indian Institute of Technology Roorkee, 
247667, Uttarakhand, India}
\address[3]{Graduate School of Informatics, Osaka Metropolitan University, 1-1 Gakuencho, Naka, Sakai, Osaka 599-8531 Japan}
\tfootnote{This work was supported by JSPS KAKENHI JP19K12023. }

\markboth
{Author \headeretal: Preparation of Papers for IEEE TRANSACTIONS and JOURNALS}
{Author \headeretal: Preparation of Papers for IEEE TRANSACTIONS and JOURNALS}

\corresp{Corresponding author: Katsufumi Inoue (e-mail: inoue@omu.ac.jp).}

\begin{abstract}
Word-level sign language recognition (WSLR) has attracted attention because it is expected to overcome the communication barrier between people with speech impairment and those who can hear.
In the WSLR problem, a method designed for action recognition has achieved the state-of-the-art accuracy.
Indeed, it sounds reasonable for an action recognition method to perform well on WSLR because sign language is regarded as an action.
However, a careful evaluation of the tasks reveals that the tasks of action recognition and WSLR are inherently different.
Hence, in this paper, we propose a novel WSLR method that takes into account information specifically useful for the WSLR problem.
We realize it as a multi-stream neural network (MSNN), which consist of three streams: 
1) base stream, 2) local image stream, and 3) skeleton stream. 
Each stream is designed to handle different types of information. 
The base stream deals with quick and detailed movements of the hands and body, 
the local image stream focuses on handshapes and facial expressions, 
and the skeleton stream captures the relative positions of the body and both hands. 
This approach allows us to combine various types of data for more comprehensive gesture analysis.
Experimental results on the WLASL and MS-ASL datasets show the effectiveness of the proposed method; it achieved an improvement of approximately 10\%--15\% in Top-1 accuracy when compared with conventional methods.
\end{abstract}

\begin{keywords}
3D convolutional neural network, Deep learning, Face, Hand, Optical flow, Skeleton, Word-level sign language recognition. 
\end{keywords}

\titlepgskip=-21pt

\maketitle

\section{Introduction}
\label{sec:introduction}
Sign language is one of the most important tools people with speech impairment use to communicate with others.
The mother tongue of people with speech impairment is often sign language because they grow up using sign language.
However, it is not well known that many people with speech impairment find it difficult to use written language.
This is because the grammar of sign language is often different from that of written language.
Thus, written language is like a foreign language for people with speech impairment; imagine a situation in which non-native English speakers (e.g., native Japanese speakers) speak English.
Therefore, people with speech impairment are comfortable using sign language.
For the same reason, mastering sign language is not always easy for people who can hear.
To overcome the communication barrier between people with speech impairment and those who can hear, automatic translation between sign language and spoken (written) language is desired.
As its core technique, word-level sign language recognition (Word-level SLR; WSLR) has been widely researched (e.g., \cite{li2020transferring, wang2014similarity, Hosain_2021_WACV, Camgoz_2020_CVPR, li2020word, vaezi2019ms-asl}).

The state-of-the-art results on WSLR~\cite{li2020word, vaezi2019ms-asl} were achieved using a recognition method proposed for action recognition.
The action recognition method, called the I3D network~\cite{joao2017i3d}, extracts spatiotemporal features from sequential images for recognition.
Indeed, sign language is regarded as an action.
Hence, one might think that it is reasonable for an action recognition method to perform well on WSLR.
However, a closer look at the tasks reveals that the WSLR problem is not a subset of the action recognition problem.
The action recognition task is to distinguish different actions such as standing, running, and punching.
The body movements of different actions are quite different.
By contrast, words of sign language are often similar to each other.
Taking the example of punching (i.e., boxing), the WSLR problem is like distinguishing punches such as jabs, straights, hooks, and uppercuts. 
That is, we can consider that WSLR is one of fine-grained action recognition tasks, 
and this is one of challenges of WSLR compared to the general action recognition tasks. 
Besides, for the general action recognition, the surrounding information such as background environments, objects manipulated by an actor, etc., is often useful to recognize the general actions. 
On the other hand, for WSLR, such information may not help us to recognize sign language words accurately and we need to recognize them from the only actions by signers. 
This is another challenge WSLR compared to the general action recognition.

\begin{figure}[t]
  \centering
    \includegraphics[width=\hsize]{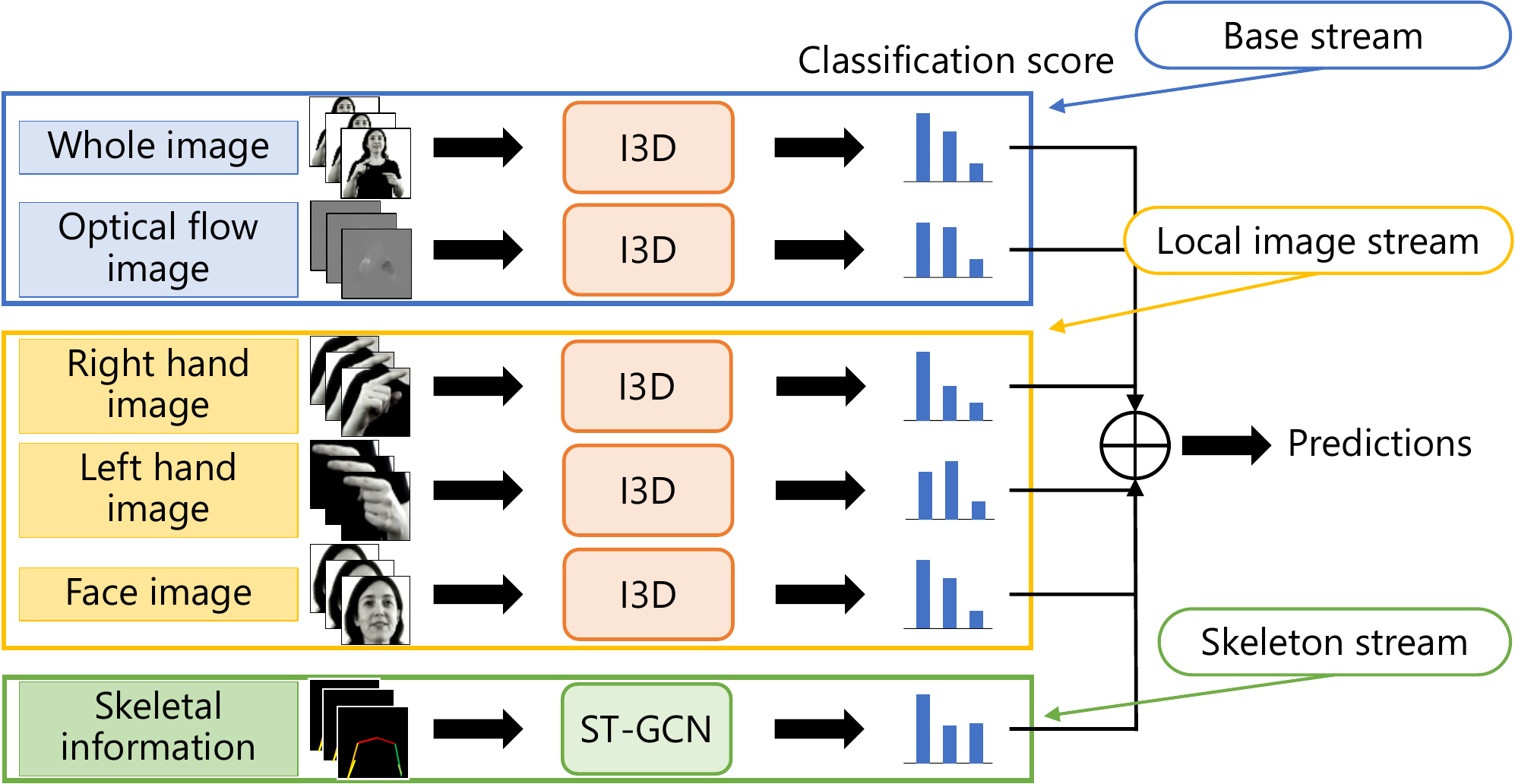}
    \caption{Overview of the proposed method.
    The proposed multi-stream neural network (MSNN) consists of three streams: 1) a base stream, 2) local image stream, and 3) skeleton stream. Each stream is trained separately, and the recognition scores extracted from each stream are averaged to obtain the final recognition result. }
    \label{fig:overview}
\end{figure}

To distinguish the actions of sign language, we observed them and found the following.
First, a word in sign language consists of consecutive actions, each of which may be shared with other words.
Second, some signs represent different words with a similar gesture; for example, only the handshapes are different.
These issues do not appear in the action recognition problem.
Therefore, in WSLR, we need to capture more detailed information useful for distinguishing sign language gestures.
However, the conventional researches mainly focus on global information 
such as appearance information extracted from the upper body of the signer. 
Although some researches focus on the local information such as handshapes, 
these basically utilize single local information. 
Compared with them, we also focus on local information and combine multiple information for WSLR. 
To do so, the paper written by Adria \etal~\cite{adria2018zoom} is suggestive.
In face recognition, they demonstrated that using face images whose eye regions are magnified improves the recognition accuracy.
This is because eye regions are more informative for distinguishing faces than other parts.
Adria \etal revealed that focusing on the informative parts improves accuracy.
In SLR, the informative factors include quick and detailed hand and body movements, handshapes, facial expression, and the relative positions of the body and both hands.

Hence, in this paper, we propose a novel WSLR method that can exploit the informative factors useful for distinguishing the actions of sign language.
As shown in \cref{fig:overview}, we realize it as an MSNN to combine various information obtained from multiple information sources.
The streams are categorized into three types: the base stream, local image stream, and skeleton stream.
The first stream, the base stream, deals with the global information of the upper body of the signer.
It consists of two sub-streams.
One uses the appearance information extracted from whole frame images in the input sign language video, which is used in the conventional methods.
The other uses the optical flow information to capture the signer's dynamic gesture movements in consecutive frame images.
The second stream, the local image stream, deals with the magnified images of both hands and face, which express the local information of handshapes and facial expressions.
Its three sub-streams are dedicated to the magnified images of the left hand, right hand, and face.
The third stream, the skeleton stream, captures the relative positions of the body and both hands in the form of a skeleton.
Because the skeleton focuses only on the signer's body, this information should reduce the effect of background noise and the signer's appearance.
The base and local image streams use I3D networks, whereas the skeleton stream uses a spatiotemporal graph convolutional network (ST-GCN)~\cite{yan2018stgcn} to capture the relative positions.

The contributions of our paper are as follows.
\begin{itemize}
\item We point out the inherent differences between action recognition and WSLR.

\item We present a novel MSNN consisting of three streams that exploits local hand and face information as well as the relative positions of the body and both hands for WSLR.

\item Our approach has achieved improvements of approximately 15\% and 10\% in Top-1 accuracy when compared with previous state-of-the-art methods on the large-scale WSLR datasets WLASL~\cite{li2020word} and MS-ASL~\cite{vaezi2019ms-asl}, respectively, although the results on the MS-ASL dataset are indicative because of the unavailability of the original data.
\end{itemize}

The remainder of this paper is organized as follows. In \cref{related}, some relevant studies in the field of SLR are discussed.
The proposed framework is detailed in \cref{proposed_method}.
\cref{experiment} describes the datasets, implementation details, and experimental results.
\cref{conclusion} concludes the paper.

\section{Related Work}
\label{related}
SLR models are discussed in depth by Rastgoo \etal~\cite{rastgoo2020sign}.
Categorically, SLR is divided into two main scenarios.
They are isolated SLR~\cite{li2020transferring, wang2014similarity} and continuous SLR~\cite{cui2019deep, evangelidis2014continuous, koller2015continuous,Wei_IEEETCSVT2021}.
Although earlier models are confined to isolated SLR, later models have been extended to continuous SLR.
In isolated SLR, gloss information is temporally well segmented and the target is to recognize a word or expression.
In contrast, continuous SLR does not use detailed information about a word or expression but uses only sentence-level annotations as a whole.
Compared with isolated SLR, the task of continuous SLR is more difficult because it aims to recognize glosses in the scenarios based on sentence-level annotations.

Traditionally, handcrafted features~\cite{wang2014similarity, buehler2009learning, pfister2014domain, evangelidis2014continuous, koller2015continuous, monnier2014multi} were used to capture spatiotemporal representations for SLR.
Because sign language involves the use of different parts of the body, such as the head, fingers, body, arm, hand, and facial expression~\cite{cheok2019review}, handcrafted features like image pixel intensity~\cite{wang2014similarity}, local binary patterns (LBP)~\cite{wang2014similarity}, histogram of oriented gradients (HOG)~\cite{buehler2009learning, pfister2014domain}, HOG-3D~\cite{koller2015continuous}, and motion trajectories or velocities~\cite{evangelidis2014continuous, koller2015continuous, monnier2014multi} were used for SLR.
For example, Koller \etal~\cite{koller2015continuous} exploited the importance of tracking for SLR with respect to the hands and facial landmarks and presented a statistical recognition approach that could perform large vocabulary continuous SLR across different signers.
Evangelidis \etal~\cite{evangelidis2014continuous} exploited dynamic programming (DP) for continuous gesture recognition
from articulated poses.
Pfister \etal~\cite{pfister2013large} developed a multiple instance learning method based on efficient discriminative search that could automatically learn a large number of signs from sign
language-interpreted TV broadcasts by employing the supervisory information available in the subtitles of the broadcasts as labels.

The recent trend of employing deep neural networks that use convolutional neural networks (CNNs) for feature extraction has substantially boosted the performance of SLR systems~\cite{pigou2014sign, molchanov2016online, neverova2014multi, huang2018attention,Cheng_IEEETCSVT2016, Li_IEEETCSVT2018, SanchezRiera_IEEETCSVT2018}.
Initially, two-dimensional (2D) CNNs were used to learn visual feature representation for SLR, which was later extended by introducing 3D CNNs, which are better able to learn spatiotemporal feature representation.
For example, Molchanov \etal~\cite{molchanov2016online} employed a recurrent 3D CNN for spatiotemporal feature extraction from the depth, color, and optical flow data in video streams to perform simultaneous detection and classification of dynamic hand gestures.
Neverova \etal~\cite{neverova2014multi} proposed a deep learning-based multi-scale and multimodal framework along with a progressive learning procedure for gesture localization, detection, and recognition.
Huang \etal~\cite{huang2018attention} proposed an attention-based 3D CNN for SLR that uses spatial and temporal attentions.
There are researches that used depth sensors~\cite{Cheng_IEEETCSVT2016, Li_IEEETCSVT2018, SanchezRiera_IEEETCSVT2018}.

By contrast, other studies on SLR have focused on temporal models~\cite{gweth2012enhanced, koller2016deeps, wu2016deep,pigou2018beyond, cui2019deep}, where the target is to learn the correspondence between sequential representations and gloss labels.
Gweth \etal~\cite{gweth2012enhanced} presented a Gaussian hidden Markov model (HMM) that uses appearance-based features from the original images and features derived from a multilayer perceptron for automatic SLR.
Koller \etal~\cite{koller2018deep} introduced the end-to-end embedding of a CNN into an HMM to combine the sequence modeling capabilities of HMMs with the strong discriminative capabilities of CNNs.
Wu \etal~\cite{wu2016deep} presented a semi-supervised hierarchical deep dynamic neural framework based on an HMM for multimodal gesture recognition, in which depth, skeleton joint information, and RGB images are treated as multimodal input observations.
Although HMMs were the most widely used temporal model in the past, the success of recurrent neural networks (RNNs) in the context of text recognition~\cite{cheng2017focusing, cheng2018aon, li2019show}, speech recognition~\cite{graves2014towards}, and machine translation~\cite{bahdanau2014neural, sutskever2014sequence} has motivated SLR researchers to explore the application of RNNs in SLR.
Considering this direction, Pu \etal~\cite{pu2019iterative} presented an RNN based alignment network with iterative optimization for weakly supervised SLR.
Pigou \etal~\cite{pigou2018beyond} developed an end-to-end neural framework composed of an RNN and temporal convolutions.
For continuous SLR, Cui \etal~\cite{cui2019deep} presented a deep CNN with stacked temporal fusion layers for feature extraction and bi-directional RCNNs for sequence learning.

Recently, \cite{yan2018spatial} introduced graph-based modeling, in which a graph convolutional network (GCN) is applied to skeleton data.
This graph-based approach has been enhanced and employed by others~\cite{li2019actional,shi2019skeleton,shi2019two,shi2020skeleton,si2019attention,cheng2020decoupling,jiang2021skeleton}.
Specifically, \cite{li2019actional} designed the actional-structural graph convolutional networks (AS-GCN) model to increase the recognition rate.
The model concentrates on the latent joints in the skeleton structure.
\cite{shi2019two} proposed a two-stream technique to boost the performance of the recognizer.
This work was further explored by \cite{shi2020skeleton}, and the streams were extended from two to four.
\cite{cheng2020decoupling} proposed a decoupling GCN to enhance the capacity of the model without increasing the computational cost.
A ResNet-based GCN architecture was proposed in \cite{song2020stronger} to enhance the model capability with less computational cost.
However, the skeleton-based approach remains under-explored.
The trials~\cite{de2019spatial} to apply ST-GCN straightforwardly to SLR were unsuccessful and yielded lower recognition accuracy than some of the handcrafted features methods.
Additionally, Al-Hammadi et al.~\cite{Al-Hammadi_3DGCN_Sensors2022} utilized an ST-GCN-like graph convolutional neural network and MediaPipe~\cite{Lugaresi_MediaPipe2019} to extract hand and body joints for representing the skeletal information of a signer. 
Although their method effectively captures local information, the absence of appearance information in local parts leads to a significant reduction in recognition accuracy when dealing with larger datasets. 
In other words, capturing detailed visual information of local parts is crucial for accurately differentiating sign language gestures in WSLR. 
To address this issue, we introduce a ``Local Stream'' to provide appearance information of local regions in addition to the skeletal information. 
By integrating the local, skeletal, and whole-body information, we aim to enhance the accuracy of WSLR in this research.


\section{Method}
\label{proposed_method}


Our idea is to combine multiple types of information with an MSNN.
An overview of the proposed method is shown in \cref{fig:overview}. 
In the proposed method, the MSNN is divided into the following three streams: 
1) the base stream, which deals with global information such as the global appearance and optical flow information,
2) the local image stream, which handles local information such as handshapes and facial expression, 
and 3) the skeleton stream, which handles the relative positions of the body and both hands.
The proposed method's characteristic points are that each stream is separately trained, and the classification scores obtained from each stream are averaged in the test phase. 
Note that we empirically decide to employ the simple averaging strategy for the integration of the information extracted from each steam via a preliminary experiment~\footnote{In this experiment, we compared the three strategies: 
1) classification scores obtained from each stream were simply averaged (proposed strategy), 
2) we stacked all input images by increasing the number of input channels and input the stacked images into I3D (early fusion strategy), 
and 3) we concatenated all the features extracted from each I3D feature extraction part and then fed them into 2-layer fully-connected neural network (late fusion strategy). 
From this experiment with WLASL dataset, we confirmed that the latter two strategies significantly reduced the recognition accuracies. 
In contrast, the proposed strategy achieved higher recognition accuracy.}, which investigates the fusion strategy. 
After these processes have been performed, the word having the highest score is considered to be the recognition result.
Each stream is explained in more detail in the following subsections.
Note that, since our research employs publicly available datasets, we do not obtain informed consent from subjects. 




\subsection{Base Stream}

\begin{figure}[t]
  \centering

      \begin{minipage}{\hsize}
        \centering
          \includegraphics[width=\hsize]{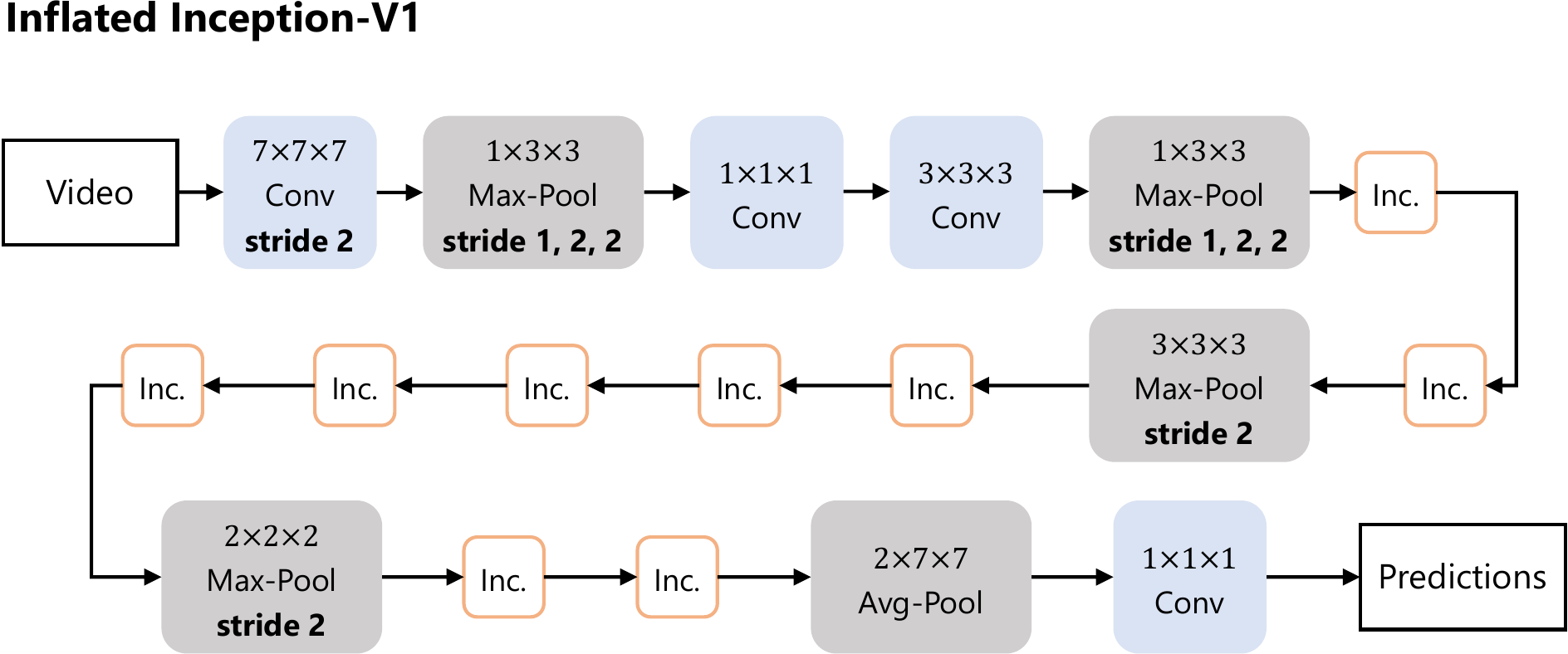}
          \subcaption{}
      \end{minipage}
      
      \vspace{1.5em}

      \begin{minipage}{\hsize}
        \centering
          \includegraphics[width=0.75\hsize]{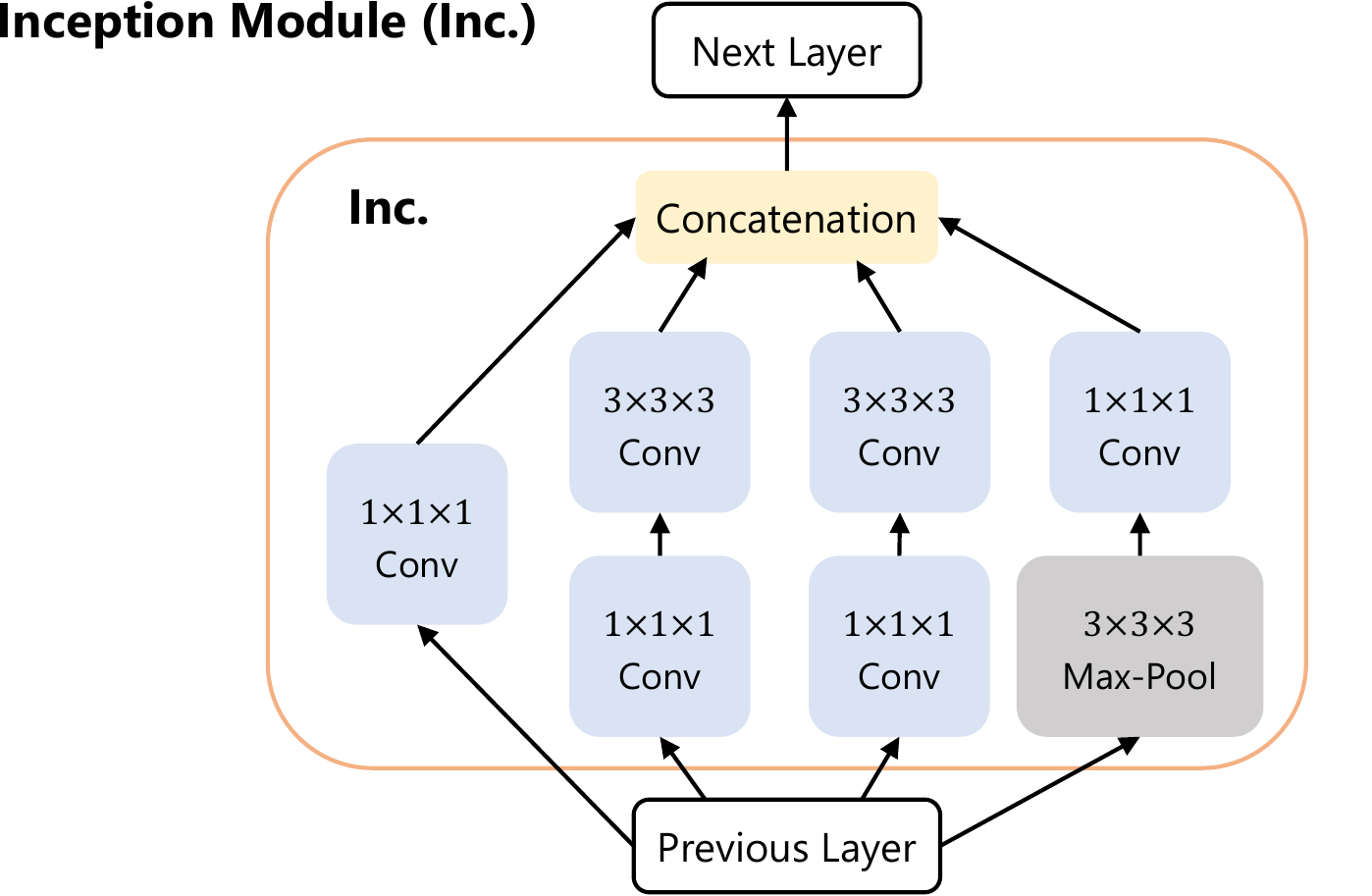}
          ~~~~~~~
          \subcaption{}
      \end{minipage}

    \caption{Overview of I3D network architecture. (a)~The whole I3D network architecture and (b)~Details of the inception submodule~\cite{joao2017i3d} in I3D.}
    \label{fig:i3d}
  
\end{figure}
As shown in \cref{fig:overview}, the base stream consists of two sub-streams to process the global information of the upper body.
The first sub-stream is for the appearance information extracted from whole frame images in the input sign language video;
it is solely used in the conventional methods~\cite{li2020word, vaezi2019ms-asl} to achieve state-of-the-art accuracies.
Hence, the first stream is considered to be the baseline of the proposed method.
In the second sub-stream, optical flow information is used to capture the signer's dynamic gesture movements in the consecutive frame images. 
In an action recognition study, Carreira and Zisserman~\cite{joao2017i3d} reported that optical flow information allows us to improve recognition accuracy by combining sequential appearance information in a two-stream structure network~\cite{joao2017i3d}. 
Inspired by this report, 
in this research, 
this two-stream I3D is also considered to be a baseline of the proposed method. 

Before explaining the concrete processes of the base stream, we briefly introduce I3D.
\cref{fig:i3d} shows an overview of the I3D. 
The characteristics of I3D is to employ many inception modules 
depicted as ``Inc.'' in~\cref{fig:i3d}~(a). 
Besides, as shown in~\cref{fig:i3d}~(b), I3D expands the 2D filters as pooling kernels of the inception network~\cite{sze2015incep} trained on the ImageNet dataset~\cite{olga2015imagenet} into 3D ones, and these inflated 3D filters are incorporated into the network.
In this research, these filters are fine-tuned with the Kinetics dataset~\cite{joao2017i3d} to extract better temporal and spatial features from an input video.
In the stream for appearance information, $M$ consecutive frame images of the input sign language video are given as input to the I3D.
In contrast, in the stream for optical flow information, $M$ optical flow images calculated from $M+1$ consecutive frame images using TV-L1 algorithm~\cite{zach2007tvl1} are input to the I3D.
Like \cite{tu2018mscnn, bilen2018multi, zang2018multi}, each stream is trained separately and outputs the classification scores of sign language words.
These scores are combined for use in in the final recognition process. 



\subsection{Local Image Stream}

\begin{figure}[t]
  \centering
    \includegraphics[width=\hsize]{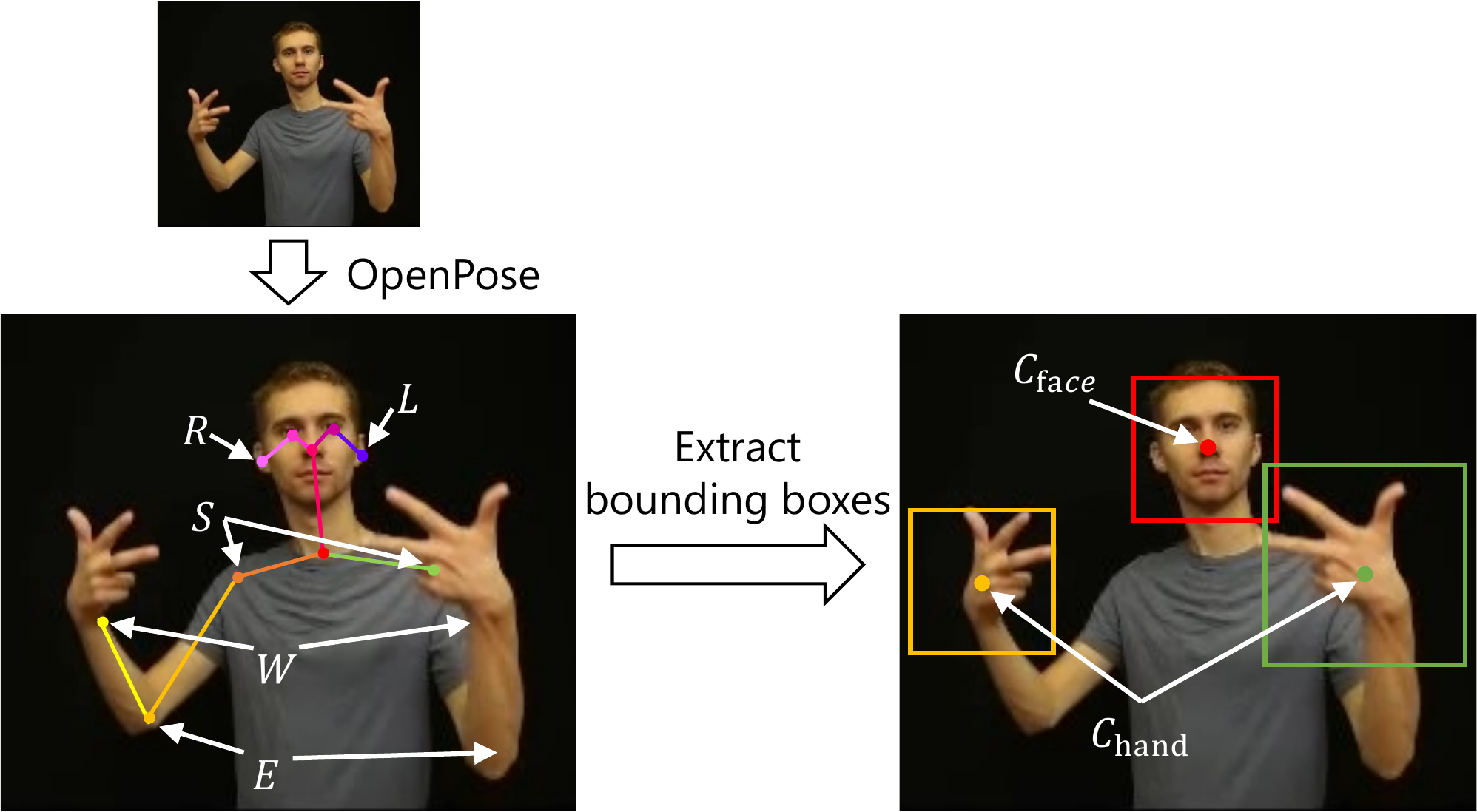}
    \caption{Example of bounding box extraction for the face and both hands.
    These bounding boxes are extracted based on the skeletal points detected with OpenPose: 
    both hand regions are based on shoulder point $S$, elbow point $E$, and wrist point $W$, and the face region is based on the left ear point $L$ and right ear point $R$. 
    (The images are provided from the WSASL dataset~\cite{li2020word})}
    \label{fig:bbox}
\end{figure}

The local image stream captures local information that cannot be sufficiently obtained from the upper body information. 
As previously mentioned, a study by \cite{adria2018zoom} found that combining local and global information can improve face recognition accuracy. 
In this study, the local parts of a face image, such as the eyes, were distorted and enlarged to emphasize the local regions. 
Interestingly, this simple image processing technique was effective in capturing the local information of the face. 
Inspired by these findings, we believe that local information has the potential to improve the accuracy of WSLR. 
In WSLR, the shape of the hand is one of the most critical factors for local information.
If the handshapes of gestures are different, even though the hand movements are the same, the gestures represent different sign words. 
Therefore, to improve the recognition accuracy of WSLR, the precise handshape should be used. 
In addition to the handshape information, facial information is also an important factor because Deaf people read signs by capturing the signer's facial expression and mouth movements. 
Thus, the face region is also extracted and used for understanding the facial information accurately. 
To understand this information precisely, the handshape and facial information of each sign word are separately learned from the appearance information of the upper body of the signer. 
Note that, in contrast to the approach in \cite{adria2018zoom}, we do not distort and enlarge the local regions; instead, we crop them from the input frame image to concentrate on these areas.
To realize this in the local image stream, an MSNN with three sub-streams is constructed and used for the left hand, right hand, and face regions.

The concrete processes of the local image stream are explained as follows. 
First, the signer's skeletal points are detected with OpenPose~\cite{cao2017openpose}, a method to obtain the coordinates and likelihood of the skeletal points of the signer. 
From the positions of these skeletal points, all hand and face regions are extracted with bounding boxes, as shown in \cref{fig:bbox}. 
To extract both hand regions, it is assumed that the hand is on the elbow's extension to the wrist. 
From this assumption, the bounding boxes of the hands are obtained and calculated by
\begin{align}
    \label{eq:EChand}
    \overrightarrow{EC_\textrm{hand}} &= 1.33 \times \overrightarrow{EW} \\
    \label{eq:WHhand}
    w_\textrm{hand} = h_\textrm{hand} &= 1.2 \times \textrm{max}(|\overrightarrow{EW}|, 0.9 \times |\overrightarrow{SE}|),
\end{align}
where $C_{\mathrm{hand}}$ is the central point of the bounding box and $S$, $E$, and $W$ are the shoulder, elbow, and wrist points, respectively. 
Further, $w_{\mathrm{hand}}$ and $h_{\mathrm{hand}}$ are the width and height of the bounding box, respectively.
The arrows above the variables represent vectors. 
In this research, all constant numbers in \cref{eq:EChand,eq:WHhand} were empirically determined using the WLASL~\cite{li2020word} dataset to ensure that the hand regions are included in the bounding boxes.
We call the sub-streams for the left and right hands the \textit{left-hand stream} and \textit{right-hand stream}, respectively.
In contrast, for the bounding box extraction of the face region, 
we use the positions of the left and right ears detected by OpenPose;
they are calculated using the following equations based on the extracted positions: 
\begin{align}
    \label{eq:RCface}
    \overrightarrow{RC_\textrm{face}} &= \frac{1}{2} \times \overrightarrow{RL} \\
    \label{eq:WHface}
    w_\textrm{face} = h_\textrm{face} &= 1.5 \times |\overrightarrow{RL}|,
\end{align}
where $L$ and $R$ are the left and right ear points, respectively. 
Similar to the constants for the bounding boxes of the hand region, the constants in \cref{eq:RCface,eq:WHface} were also determined empirically on the above dataset to ensure that the face region is included in the bounding box.
The local images were cropped from the whole image using their bounding boxes and resized to $224 \times 224$ pixels.
Using the cropped images, the sub-stream networks were trained separately.
I3D networks were also used for training the handshape and facial information.
The outputs of the sub-streams are the classification scores of the sign language words.
Similar to the scores obtained by the base stream, each score is used for the final recognition process.

\begin{figure}[t]
  \centering
    \includegraphics[width=\hsize]{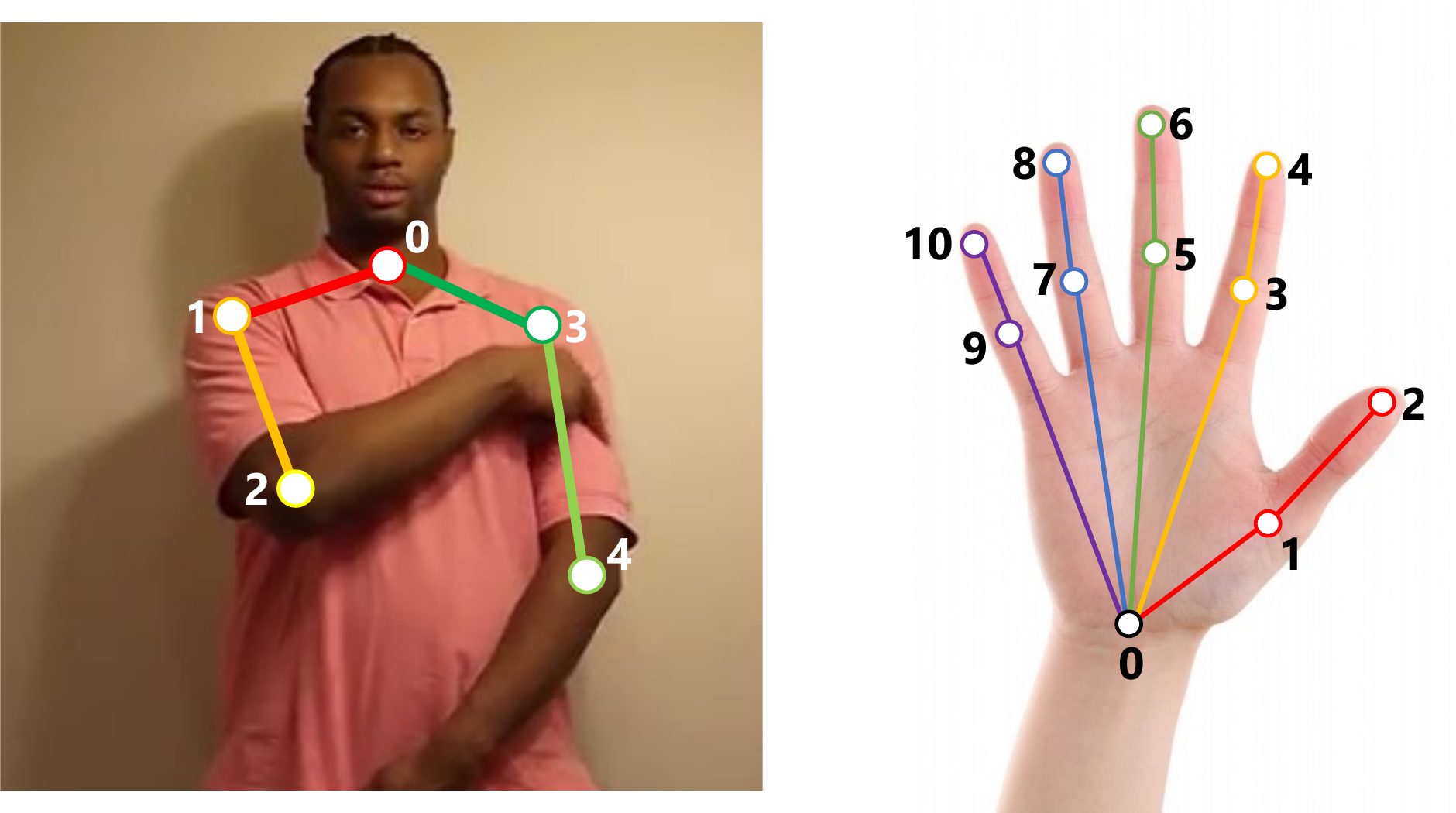}
    \caption{27 keypoints inputted to ST-GCN in the skeleton stream. Five keypoints refer to the body, and 11 keypoints refer to each hand. (The left image is from \cite{vaezi2019ms-asl}.)}
    \label{fig:keypoints}
\end{figure}


\subsection{Skeleton Stream}

\begin{figure}[t]
  \centering
    \includegraphics[width=\hsize]{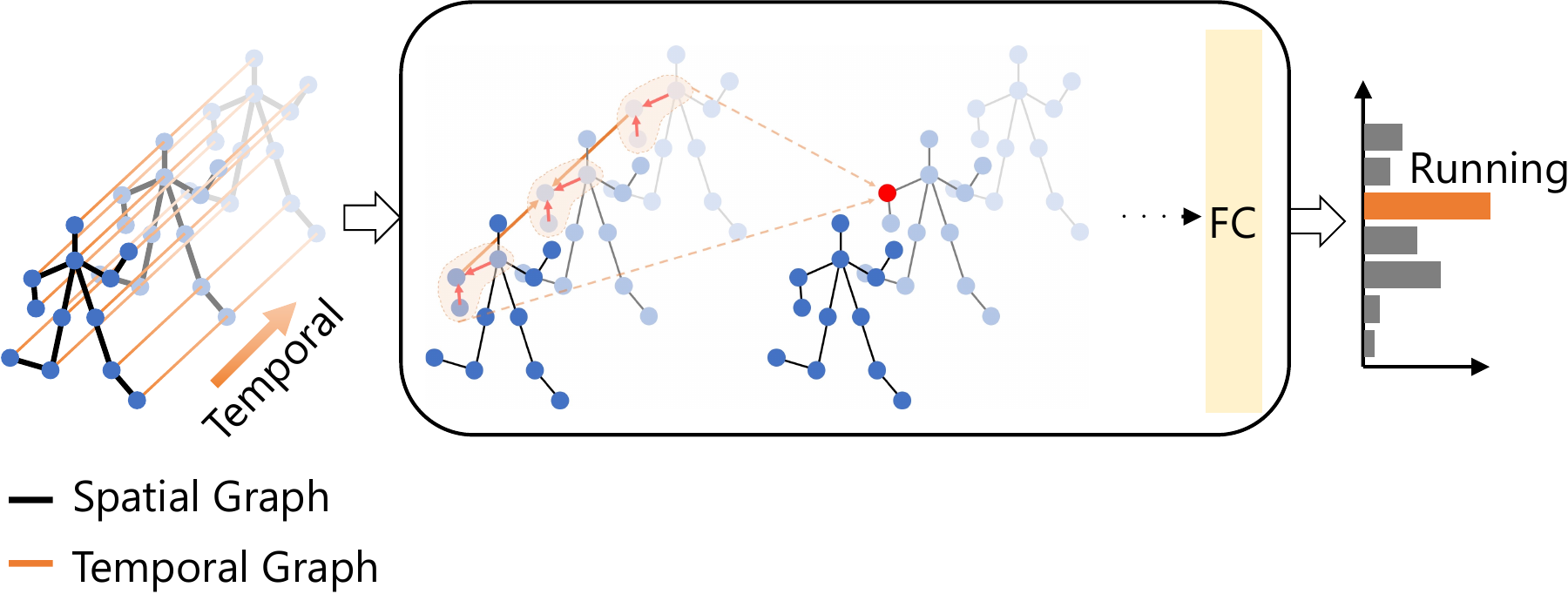}
    \caption{Overview of the ST-GCN architecture~\cite{yan2018stgcn}.}
    \label{fig:stgcn}
\end{figure}

The skeleton stream captures the signer's spatiotemporal skeletal information.
In sign language, the signer often points to various body parts such as an eye or mouth to distinguish different sign words.
That is, the meaning of sign words varies depending on where the signer is pointing.
Therefore, to classify these sign words, the relative positions of the body and both hands need to be understood precisely.
Moreover, to understand the differences in the gestures, changes in the skeleton in consecutive frame images are also checked.
In contrast to the base and local image streams, which use I3D, we use an ST-GCN~\cite{yan2018stgcn} for the skeleton stream to obtain such spatiotemporal information (relative positions and skeletal changes) for WSLR.
It has also been used for action recognition using skeletal data and achieved high recognition accuracy.

In the following, we explain the concrete process of the skeleton stream. 
First, the skeleton information from each frame image is extracted using OpenPose~\cite{cao2017openpose}, as illustrated in \cref{fig:keypoints}. 
In this study, 27 keypoints are extracted from each frame image;
five and 11 keypoints are extracted from the body region and each hand region, respectively.
After keypoint extraction, all the 2D coordinates of each keypoint are extracted from a frame image to obtain the concatenated vector $\bm{P}_\mathrm{skel}$ as the skeletal feature as follows: 
\begin{align}
    \bm{P}_\mathrm{skel} = (\bm{p}_0^b, \dots, \bm{p}_4^b, \bm{p}_0^{lh}, \dots, \bm{p}_{11}^{lh}, 
    \bm{p}_0^{rh}, \dots, \bm{p}_{11}^{rh}),
    \label{eq:skelfeature}
\end{align}
where $\bm{p}_i^b$, $\bm{p}_i^{lh}$, and $\bm{p}_i^{rh}$ are the coordinates of the $i$-th keypoints of the body, left hand, and right hand, respectively. 
Note that, if OpenPose misses a keypoint detection, we set the coordinates of that keypoint to $(0, 0)$.
Then, the features extracted from the consecutive frame images are input to the ST-GCN.
An overview of the ST-GCN is shown in \cref{fig:stgcn}.
In an ST-GCN, the skeletal data are used to construct two graph structures.
One is a spatial graph that focuses on the spatial relationship among the keypoints extracted from a frame image.
The other is a temporal graph that focuses on the temporal change by connecting the same keypoints between frames.
These graphs allow the classification model to learn both the spatial and temporal features of skeletal keypoints simultaneously.
The model outputs the classification scores of the sign language words using these mechanisms. These scores are used for the final recognition process.




\subsection{Algorithmic Procedure}
\begin{figure}[t]
    \begin{algorithm}[H]
        \caption{MSNN}
        \label{alg:pseudo}
        \begin{algorithmic}[1]
            \Require{Images $\bm{I}$}
            \Require{Optical Flow Images $\bm{F}$}
            \Require{Stream List $\bm{S}_{\mathrm{list}} \leftarrow \{S_{\mathrm{wh}}, S_{\mathrm{fl}}, S_{\mathrm{rh}}, S_{\mathrm{lh}}, S_{\mathrm{fa}}, S_{\mathrm{sk}}\}$}
            \Require{Input Map $\bm{I}^m$}
            \Require{Classifier Map $\bm{C}^m \leftarrow \{$I3D, ST-GCN$\}$}
            \Ensure{Word Class $c$}
            \Function {MSNN}{ $I$ }
                \State{SET\_INPUT\_MAP( $\bm{I}^m, \bm{I}, \bm{F}$ )}
                \ForAll{ $stream \gets \bm{S}_\mathrm{list}$}
                    \State{$input \gets \bm{I}^m(stream)$}
                    \State{$classifier \gets \bm{C}^m(stream)$}
                    \State{$Prob \gets classifier(input)$}
                    \State{\textbf{Push} ($prob\_all, prob$)}
                \EndFor
                \State{$prob\_final \gets $ AVERAGE($prob\_all$)}
                \State{$c \gets$ ARGMAX\_CLASS($prob\_final$)}
            \EndFunction
            \Function{SET\_IMPUT\_MAP}{ $\bm{I}^m, \bm{I}, \bm{F}$ }
                \State{\textbf{Push} ($\bm{I}^m, \bm{I}$) }
                \State{\textbf{Push} ($\bm{I}^m, \bm{F}$) }
                \State{\textbf{Push} ($\bm{I}^m$, EXTRACT\_RIGTH\_HAND($\bm{I}$))}
                \State{\textbf{Push} ($\bm{I}^m$, EXTRACT\_LEFT\_HAND($\bm{I}$))}
                \State{\textbf{Push} ($\bm{I}^m$, EXTRACT\_FACE($\bm{I}$))}
                \State{\textbf{Push} ($\bm{I}^m$, CALCULATE\_SLEKETON($\bm{I}$))}
            \EndFunction
        \end{algorithmic}
    \end{algorithm}
\end{figure}

In this section, we explain the algorithmic procedure of the proposed method. 
\cref{alg:pseudo} shows the pseudo-code of the proposed method. 
As depicted in \cref{fig:overview}, the MSNN includes six streams,  
$S_\mathrm{wh}, S_\mathrm{fl}, S_\mathrm{rh}, S_\mathrm{lh}, S_\mathrm{fa}$, and $S_\mathrm{sk}$. 
These streams handle global appearance, optical flow, right hand, left hand, face, and skeleton information, respectively, and are collectively included in the set $S_{\mathrm{list}}$. 
As explained in the previous subsections, to deal with each information, 
we separately prepare input data: consecutive frame images $\bm{I}$ for $S_\mathrm{wh}$, 
optical flow images $\bm{F}$ for $S_\mathrm{fl}$, 
right and left hand images, face images, and skeleton graphs extracted from frame images $\bm{I}$ using OpenPose for $S_\mathrm{rh}, S_\mathrm{lh}$, $S_\mathrm{fa}$, and $S_\mathrm{sk}$. 
This preparation procedure is described in SET\_INPUT\_MAP function in \cref{alg:pseudo}. 
Besides, in the proposed method, ST-GCN is employed for $S_\mathrm{sk}$ and 
I3D is utilized for other streams. 
Therefore, we set classifier map $\bm{C}^m$ as shown in \cref{alg:pseudo}. 
From these settings, for each stream, the probability scores of sign language words are calculated. 
After that, these scores for each word are averaged, and then the sign language word having the maximum score is regarded as the recognition result.


\section{Experiments}\label{experiment}



\subsection{Datasets}
\label{subsec:datasets}
We used the WLASL dataset~\cite{li2020word} and MS-ASL dataset~\cite{vaezi2019ms-asl}, which are representative datasets for American sign language (ASL) recognition that include a large number of classes, signers, videos, and videos per class.
Especially, since WLASL is the largest ASL word dataset, the challenge of this dataset is scalability. 
On the other hand, the videos in MS-ASL dataset are captured from variety of viewpoints. 
Therefore, this variety of viewpoints for capturing the signers is the challenge of the MS-ASL dataset. 
In each video of these datasets, a native ASL signer or interpreter performs only one sign word.
As summarized in \cref{tab:dataset}, both datasets consist of four subsets;
the number after each subset name (e.g., ``1000'' of WLASL1000) indicates the number of classes included in the subset.
These subsets consist of Top-$K$ classes when the classes are sorted in the ascending order of the number of videos per class, where $K=\{100, 300, 1000, 2000\}$ in the WLASL dataset and $K=\{100, 200, 500, 1000\}$ in the MS-ASL dataset.
We evaluated the proposed method on all subsets of the WLASL and MS-ASL datasets.
However, some of the video links to download the MS-ASL dataset have expired, and thus we could not obtain approximately 20\%--25\% of the data in the original MS-ASL dataset.
Therefore, the results on the MS-ASL dataset should be taken as indicative.
Following the splitting rule provided by the authors of the datasets, the datasets were split so that the ratio of training, validation, and test data was 4:1:1.


\subsection{Implementation Details}
In the evaluation of the proposed method, the training and testing strategies followed the conventional researches~\cite{li2020word, vaezi2019ms-asl}.
That is, as a preprocessing step, the bounding boxes of signers in the WLASL and MS-ASL datasets were detected by YOLOv3~\cite{redmon2018yolov3} and SSD~\cite{liu2016ssd}, respectively.
Then, the bounding boxes were enlarged by a factor of $\sqrt{2}$ in the following two-step process.
First, each frame was resized so that the diagonal size of the signer's bounding box was 256 pixels.
That is, the size of the bounding box was $(256/\sqrt{2}) \times (256/\sqrt{2})$.
Second, for each frame, a $256 \times 256$ square region centered on the signer's bounding box was cropped.
In the training phase, we applied the following data augmentation.
As spatial data augmentation, a $224 \times 224$ patch was randomly cropped from the $256 \times 256$ square region.
Additionally, random horizontal flipping with a probability of 0.5 was applied to the normalized frames because the meaning of a sign does not change in ASL even if it is mirrored.
As temporal data augmentation, $M = 64$ consecutive normalized frames were randomly selected as input to all streams.
For videos shorter than 64 frames, either the first or the last frame was randomly selected, and then the videos were padded by repeatedly duplicating the selected frame. 
We trained I3D using the Adam optimizer with an initial learning rate of $10^{-3}$ and a weight decay of $10^{-7}$. 
Moreover, we trained the ST-GCN using Adam with an initial learning rate of 0.01  and a weight decay of $10^{-4}$.
All models were trained with 200 epochs on each dataset.
In the test phase, all frames of the videos were input to the model.


\subsection{Quantitative Evaluation}
As shown in \cref{tab:result}, we compared the proposed method with two baselines.
The first baseline is I3D using only whole images as input.
This method was proposed in~\cite{li2020word, vaezi2019ms-asl} and has achieved state-of-the-art results on both datasets.
The second baseline is a two-stream I3D using whole images and optical flow images as input.
In other words, the baselines consist of only the base stream in \cref{fig:overview}.
We named these two baselines \textit{Baseline1} and \textit{Baseline2}, respectively.
On top of the baselines, all combinations of with and without the local image and skeleton streams were evaluated to determine the effectiveness of these streams.
We call these models \textit{Ours1} through \textit{Ours6}.
Moreover, we compared the proposed method with other state-of-the-art methods~\footnote{Several papers~\cite{SPOTER_WACV2022, SignBERT_ICCV2021, DU_FullTransformer2022, Guo_NCA2023, Hruz_OneModel2022, Rajalakshmi_IEEE_Access2022} were proposed after the pre-print version of this paper~\cite{Maruyama_MSNN2021} was published on arXiv. Some of these works~\cite{DU_FullTransformer2022, Guo_NCA2023, Hruz_OneModel2022, Rajalakshmi_IEEE_Access2022} referenced our arXiv paper and compared their methods with ours. In this paper, we also provide a comparison with these studies and present the recognition results. } to determine the effectiveness of our method.
We also evaluated the recognition accuracy of the methods using Top-$N$ classification accuracies with $N=\{1, 5, 10\}$.

\begin{table}[t]
    \centering
    \caption{Details of the datasets. \#Classes, \#Videos, and \#Signers denote the numbers of classes, videos, and signers, respectively. Column \textit{Mean} denotes the average number of videos per class.
    Regarding the MS-ASL dataset, we could download only a part of the designated videos because some links were no longer valid. Hence, the table also shows in the parentheses the amount of the data we could download compared with the original.
    }
    \footnotesize
    \begin{tabular}{ccccc} \hline
        Subset & \#Classes & \#Videos & Mean & \#Signers \\ \hline \hline
        WLASL100~\cite{li2020word} & 100 & 2,038 & 20.4 & 97 \\
        WLASL300~\cite{li2020word} & 300 & 5,117 & 17.1 & 109\\
        WLASL1000~\cite{li2020word} & 1,000 & 13,168 & 13.2 & 116\\
        WLASL2000~\cite{li2020word} & 2,000 & 21,083 & 10.5 & 119 \\ \hline
        \multirow{2}{*}{MS-ASL100~\cite{vaezi2019ms-asl}} & \multirow{2}{*}{100} & 4,315 & 43.2 & 163 \\
         & & (-25\%) & (-25\%) & (-13.8\%) \\
        \multirow{2}{*}{MS-ASL200~\cite{vaezi2019ms-asl}} & \multirow{2}{*}{200} & 7,431 & 37.2 & 171 \\
         & & (-23.5\%) & (-23.5\%) & (-12.8\%) \\
        \multirow{2}{*}{MS-ASL500~\cite{vaezi2019ms-asl}} & \multirow{2}{*}{500} & 13909 & 27.8 & 188 \\
         & & (-22.0\%) & (-23.5\%) & (-15.3\%) \\
        \multirow{2}{*}{MS-ASL1000~\cite{vaezi2019ms-asl}} & \multirow{2}{*}{1,000} & 20,092 & 20.1 & 196 \\
         & & (-21.2\%) & (-21.2\%) & (-11.7\%) \\ \hline
    \end{tabular}
    \label{tab:dataset}
\end{table}


\subsubsection{Experimental results on WLASL dataset}
\begin{table*}[ht]
    \centering
    \renewcommand{\arraystretch}{1.3}

    \caption{Compositions of the methods and their recognition accuracies (\%).
    Columns \textit {RGB}, \textit{Flow}, \textit{Local}, and \textit{Skeleton} denote whole images, optical flow images, local image patches, and skeletal information, respectively. The four row-blocks display the experimental results of baselines, competitive methods introduced before and after the release of our pre-print~\cite{Maruyama_MSNN2021}, and our proposed methods, respectively. The bolded results in the second to fourth blocks represent the best accuracy in each block. }
    \label{tab:result}

    \begin{minipage}{\hsize}
        \centering
    \subcaption{Four subsets of WLASL dataset.}
    \label{tab:wlasl}
    \scalebox{0.8}{
    \begin{tabular}{lccccccccccccccccccc} \hline
         & & & & & \multicolumn{3}{c}{WLASL100} && \multicolumn{3}{c}{WLASL300} && \multicolumn{3}{c}{WLASL1000} && \multicolumn{3}{c}{WLASL2000}\\ \cline{6-8} \cline{10-12} \cline{14-16} \cline{18-20}
        Model & RGB & Flow & Local & Skeleton & Top 1 & Top 5 & Top 10 && Top 1 & Top 5 & Top 10 && Top 1 & Top 5 & Top 10 && Top 1 & Top 5 & Top 10 \\ \hline \hline
        Baseline1~\cite{li2020word} & \checkmark & & & & 65.89 & 84.11 & 89.92 && 56.14 & 79.94 & 86.98 && 47.33 & 76.44 & 84.33 && 32.48& 57.31 & 66.31\\
        Baseline2 & \checkmark & \checkmark & & & 77.55 & 91.25 & 94.92 && 66.96 & 87.61 & 92.03 && 56.35 & 83.03 & 88.77 && 38.67 & 68.43 & 76.39 \\ 
        \hline
        Cross-domain~\cite{li2020transferring} & \checkmark & & & & 77.52 & \textbf{91.08} & - && \textbf{68.56} & \textbf{89.52} & - && - & - & - && - & - & - \\
        Fusion-3~\cite{Hosain_2021_WACV} & \checkmark & & & \checkmark & \textbf{75.67} & 86.00 & \textbf{90.16} && 68.30 & 83.19 & \textbf{86.22} && \textbf{56.68} & \textbf{79.85} & \textbf{84.71} && \textbf{38.84} & \textbf{67.58} & \textbf{75.71} \\
        GCN-BERT~\cite{Tunga_WACV} & & & & \checkmark & 60.15 & 83.98 & 88.67 && 42.18 & 71.71 & 80.93 && - & - & - && - & - & - \\ \hline
        SPOTER~\cite{SPOTER_WACV2022} & & & & \checkmark & 63.18 & - & - && 43.78 & - & - && - & - & - && - & - & - \\ 
        SignBERT(H+P)~\cite{SignBERT_ICCV2021} & & & & \checkmark & 79.07 & 93.80 & - && 70.36 & 88.92 & - && - & - & - && 47.46 & 83.32 & - \\ 
        SignBERT(H+R)~\cite{SignBERT_ICCV2021} & \checkmark & & & \checkmark & 82.56 & \textbf{94.96} & - && 74.40 & 91.32 & - && - & - & - && \textbf{54.69} & \textbf{87.49} & - \\
        FullTransformer~\cite{DU_FullTransformer2022} & \checkmark & & & & 80.72 & 93.30 & 96.17 && 70.14 & \textbf{91.33} & 93.92 && 57.13 & 85.17 & 90.57 && - & - & - \\ 
        GLR~\cite{Guo_NCA2023} & \checkmark & & & & \textbf{82.56} & 94.57 & 96.90 && \textbf{73.95} & 90.87 & 94.91 && \textbf{64.45} & 86.09 & 91.36 && 51.34 & 82.74 & 89.27 \\ 
        CMA-ES~\cite{Hruz_OneModel2022} & \checkmark & \checkmark & & \checkmark & - & - & - && 73.87 & - & - && - & - & - && - & - & - \\
        hDNN-SLR~\cite{Rajalakshmi_IEEE_Access2022} & \checkmark & \checkmark & & \checkmark & - & - & \textbf{98.75} && - & - & \textbf{98.02} && - & - & \textbf{97.94} && - & - & \textbf{97.54} \\ 
        \hline
        Ours1 & \checkmark & & \checkmark & & 76.60 & 89.13 & 92.80 && 66.34 & 88.46 & 92.43 && 56.91 & 84.55 & 89.83 && 41.01 & 74.46 & 81.85 \\
        Ours2 & \checkmark & & & \checkmark & 71.07 & 90.13 & 93.42 && 65.10 & 85.49 & 90.64 && 53.75 & 80.01 & 86.56 && 37.65 & 67.61 & 75.97 \\
        Ours3 & \checkmark & & \checkmark & \checkmark & 77.48 & 92.38 & 95.72 && 69.99 & 89.91 & 93.50 && 60.85 & 86.98 & 91.41 && 45.12 & 79.17 & 85.65 \\
        Ours4 & \checkmark & \checkmark & \checkmark & & 80.38 & 93.38 & 95.97 && 73.07 & \textbf{90.85} & 94.44 && 62.76 & 88.02 & 92.32 && 45.30 & 79.63 & 85.75 \\
        Ours5 & \checkmark & \checkmark & & \checkmark & 78.05 & 91.63 & 95.42 && 69.77 & 88.61 & 92.33 && 58.68 & 84.05 & 90.25 && 41.94 & 73.16 & 81.68 \\
        Ours6 & \checkmark & \checkmark & \checkmark & \checkmark & \textbf{81.38} & \textbf{94.13} & \textbf{96.05} && \textbf{73.43} & 90.19 & \textbf{94.83} && \textbf{63.61} & \textbf{88.98} & \textbf{92.94} && \textbf{47.26} & \textbf{81.71} & \textbf{87.47} \\ 
        \hline   
    \end{tabular}
    }
    \end{minipage}

\vspace{1em}

    \begin{minipage}{\hsize}
        \centering
    \renewcommand{\arraystretch}{1.3}
    \subcaption{Four subsets of the MS-ASL dataset.
    The results of MS-ASL are indicative because of the missing training data.
    Note that the results of \cite{li2020transferring} and \cite{SignBERT_ICCV2021} not directly comparable with our results because different data were missing in each experiment.
    }
    \label{tab:msasl}
    \scalebox{0.8}{
    \begin{tabular}{lccccccccccccccccccc} \hline
         & & & & & \multicolumn{3}{c}{MS-ASL100$^*$} && \multicolumn{3}{c}{MS-ASL200$^*$} && \multicolumn{3}{c}{MS-ASL500$^*$} && \multicolumn{3}{c}{MS-ASL1000$^*$}\\ \cline{6-8} \cline{10-12} \cline{14-16} \cline{18-20}
        Model & RGB & Flow & Local & Skeleton & Top 1 & Top 5 & Top 10 && Top 1 & Top 5 & Top 10 && Top 1 & Top 5 & Top 10 && Top 1 & Top 5 & Top 10\\ \hline \hline
        Baseline1~\cite{li2020word} & \checkmark & & & & 73.69 & 91.38 & 93.87 && 69.16 & 86.26 & 90.43 && 50.27 & 75.39 & 82.29 && 35.10 & 59.61 & 68.93\\
        Baseline2 & \checkmark & \checkmark & & & 82.46 & 94.66 & 96.61 && 79.02 & 92.46 & 95.01 && 63.36 & 84.64 & 88.51 && 50.59 & 73.89 & 80.08\\ 
        \hline
        Cross-domain~\cite{li2020transferring} & \checkmark & & & & \textbf{83.04} & \textbf{93.46} & - && \textbf{80.31} & \textbf{91.82} & - && - & - & - && - & - & - \\ \hline
        SignBERT(H+P)~\cite{SignBERT_ICCV2021} & & & & \checkmark & 81.37 & 93.66 & - && 77.34 & 91.10 & - && - & - & - && 59.80 & 80.94 & - \\ 
        SignBERT(H+R)~\cite{SignBERT_ICCV2021} & \checkmark & & & \checkmark & \textbf{89.56} & \textbf{97.36} & - && \textbf{86.98} & \textbf{96.39} & - && - & - & - && \textbf{71.24} & \textbf{89.12} & - \\ 
        \hline
        Ours1 & \checkmark & & \checkmark & & 75.12 & 91.33 & 93.87 && 73.11 & 89.27 & 92.38 && 56.37 & 78.00 & 83.76 && 39.55 & 63.09 & 70.49\\
        Ours2 & \checkmark & & & \checkmark & 75.61 & 92.38 & 95.07 && 70.17 & 86.27 & 91.02 && 52.63 & 75.81 & 82.96 && 35.61 & 59.50 & 67.93\\
        Ours3 & \checkmark & & \checkmark & \checkmark & 76.69 & 92.65 & 95.35 && 73.32 & 89.42 & 92.18 && 58.60 & 79.18 & 85.28 && 40.23 & 63.96 & 70.92\\
        Ours4 & \checkmark & \checkmark & \checkmark & & 83.84 & 94.69 & 96.18 && 80.40 & \textbf{93.29} & 95.11 && 64.66 & 84.69 & \textbf{88.94} && \textbf{50.59} & \textbf{72.33} & \textbf{78.78}\\
        Ours5 & \checkmark & \checkmark & & \checkmark & \textbf{84.22} & 94.77 & 96.48 && 78.86 & 92.26 & 94.58 && 63.44 & 83.97 & 88.68 && 48.68 & 70.84 & 77.54\\
        Ours6 &\checkmark & \checkmark & \checkmark & \checkmark & 83.86 & \textbf{94.86} & \textbf{96.66} && \textbf{80.72} & 92.76 & \textbf{95.43} && \textbf{65.46} & \textbf{84.85} & \textbf{88.94} && 49.06 & 70.46 & 76.97\\ \hline
    \end{tabular}
    }
    \end{minipage}
\end{table*}

\cref{tab:wlasl} shows the experimental results on the four subsets of the WLASL dataset.
A comparison of the results of Baseline1 and Baseline2 in the table shows that the accuracy~\cite{li2020word} (i.e., the accuracy of Baseline1) was surpassed by the accuracies of Baseline2, which uses optical flow in addition to image input.
This result indicates that 1)~motion information, which is important for WSLR, is not fully learned in the model of Baseline1, which considers only consecutive video frames, but 2)~inputting both consecutive video frames and optical flow motion images to the model leads to successful training in Baseline2.
A similar result has also been demonstrated by~\cite{joao2017i3d} in the field of action recognition research.
Ours6 with the optical flow, local image, and skeleton streams achieved the higher Top-1 accuracy on all subsets of the WLASL dataset compared with Baselines 1 and 2.
%

We investigated the effects of introducing the local image and skeleton streams individually.
First, the recognition accuracy of the models that with and without the local image stream (e.g., Baseline1 vs Ours1 and Baseline2 vs Ours4) was explored.
We observed that the model using the local image stream outperformed those not using it; for example, in the comparison between Baseline1 and Ours1, Top-1 accuracy on the WLASL100 dataset was improved by 10.71\%, from 65.89\% to 76.60\%.
This result indicates that the proposed local image stream is effective for SLR.
Second, to verify the effectiveness of the skeleton stream, we compared the models with and without it by, for example, comparisons between Baseline1 and Ours2 and between Baseline2 and Ours5.
On all subsets, the accuracy was improved when the skeleton stream was employed, although not as significantly as when the local image stream was used; for example, in the comparison between Baseline1 and Ours2, Top-1 accuracy on the WLASL100 dataset was improved by 5.18\%, from 65.89\% to 71.07\%.
These results confirm that the skeleton stream improves SLR accuracy.

\begin{figure}[t]
  \centering
    \includegraphics[width=\hsize]{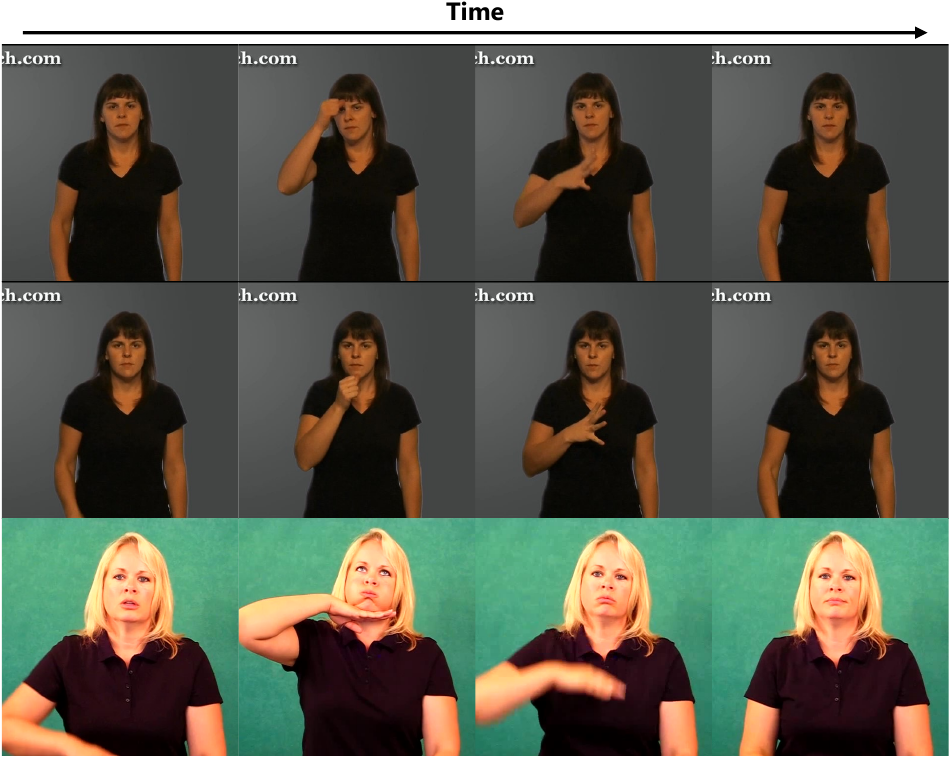}
  \caption{Example sequences of signs that represent different words with similar gestures: from the top to bottom rows, ``man,'' ``woman,'' and ``full.''}
  \label{fig:man}
\end{figure}

We next discuss how the introduction of the proposed local image and skeleton streams allows us to improve the recognition accuracy.
In particular, we focus on the word \textit{man}, for which the contribution of the proposed streams was evident.
The experimental results on the WLASL100 dataset confirmed Top-1 accuracy for \textit{man} with Baseline2 was 0\%.
However, the accuracy of Ours6, which is Baseline2 with the local image and skeleton streams, increased to 100\%.
In the results of Baseline2, \textit{man} was erroneously recognized as \textit{woman} and \textit{full}.
As shown in \cref{fig:man}, the signs for \textit{man}, \textit{woman}, and \textit{full} involve a similar series of gestures: bringing one hand up to the face and then bringing it down.
In other words, it is difficult to recognize words that contain similar gestures with a model like Baseline2, which uses only the appearance information of the whole image.
The fact that Ours6 successfully recognized these words indicates that the local image and skeleton streams are effective for WSLR, especially recognizing signs that contain similar hand and body movements.

Furthermore, by comparing our proposed method with other competitive methods, 
our method achieved better recognition accuracy compared to some image and pose-based methods~\cite{li2020transferring,Hosain_2021_WACV,Tunga_WACV}, 
and competitive recognition accuracy compared to the state-of-the-art methods~\cite{SPOTER_WACV2022,SignBERT_ICCV2021,DU_FullTransformer2022,Guo_NCA2023,Hruz_OneModel2022,Rajalakshmi_IEEE_Access2022}, including transformer-based methods~\cite{SPOTER_WACV2022, SignBERT_ICCV2021, DU_FullTransformer2022, Hruz_OneModel2022}. 
\cite{li2020transferring} extracted image features using I3D from the entire input images, without using optical flow, local image, or skeleton information. Due to these differences, as shown in \cref{tab:wlasl}, our proposed method surpassed the Top-1 accuracy by 3.86\% on WLASL100 and 4.87\% on WLASL300.
In addition, \cite{Hosain_2021_WACV} utilized I3D image features and OpenPose-based skeleton features, but it did not leverage optical flow and local image information, resulting in Ours6 outperforming it.
Moreover, Ours6, which combined three streams, allowed us to outperform some transformer-based methods~\cite{Tunga_WACV,DU_FullTransformer2022,SPOTER_WACV2022}.
These results confirmed the effectiveness of combining the three streams for WSLR.
However, it should be noted that SignBERT~\cite{SignBERT_ICCV2021}, GLR~\cite{Guo_NCA2023}, CMA-ES~\cite{Hruz_OneModel2022}, and hDNN-SLR~\cite{Rajalakshmi_IEEE_Access2022} were proposed after our pre-print version of this paper was made available, 
and these methods outperformed Ours6. 
Besides, Rajalakshmi et al. reported that HNN-ISLR~\cite{Rajalakshmi_ACM_TRAS2022} outperformed Ours6 on all subsets of the WLASL. 
This method is originally proposed for isolated Indian and Russian SLR and employs a similar approach of ours, 
i.e., multiple information such as hand, face, and pose are combined to recognize sign language words. 
The main differences between this method and ours are the feature fusion strategy and classification network. 
HNN-ISLR employes ealry-fusion strategy to combine the multiple information extracted from various cues, whereas our strategy belongs to late-fusion. 
On the other hand, for the latter difference, HNN-ISLR employes BiLSTM and fully-connected network, whereas our proposed method utilizes I3D and ST-GCN. 
From these differences, further investigation of feature fusion strategy is necessary to improve the recognition accuracy of our proposed method, 
and the investigation of more suitable classification network for WSLR is also one of our future work. 
Nonetheless, our proposed method still achieved competitive recognition accuracy compared to these state-of-the-art methods.

\subsubsection{Experimental results on MS-ASL dataset}
\cref{tab:msasl} shows the experimental results on four datasets of the MS-ASL dataset.
As stated in \cref{subsec:datasets}, the results should be taken as indicative because of the missing training data.
Overall, the experimental results of the MS-ASL dataset have a similar tendency to those of the WLASL dataset.
By adding the optical flow to Baseline1, the accuracy of Baseline2 surpassed that of Baseline1.
Ours6, with the optical flow, local image stream, and skeleton stream, achieved the highest Top-1 accuracy on the MS-ASL200 and MS-ASL500 datasets.
On the MS-ASL100 dataset, Ours6 was the second best after Ours5, with the optical flow and skeleton streams.
On the MS-ASL1000 dataset, Ours6 was the third best after Baseline2, with the optical flow, and Ours4, with optical flow and local image stream.
In contrast, Ours6 achieved the highest scores in Top-5 and Top-10 accuracies five out of times among the six times on the MS-ASL100, MS-ASL200, and MS-ASL500 datasets.
%

\begin{figure}[t]
  \centering

      \begin{minipage}[b]{0.49\hsize}
        \centering
          \includegraphics[width=\hsize]{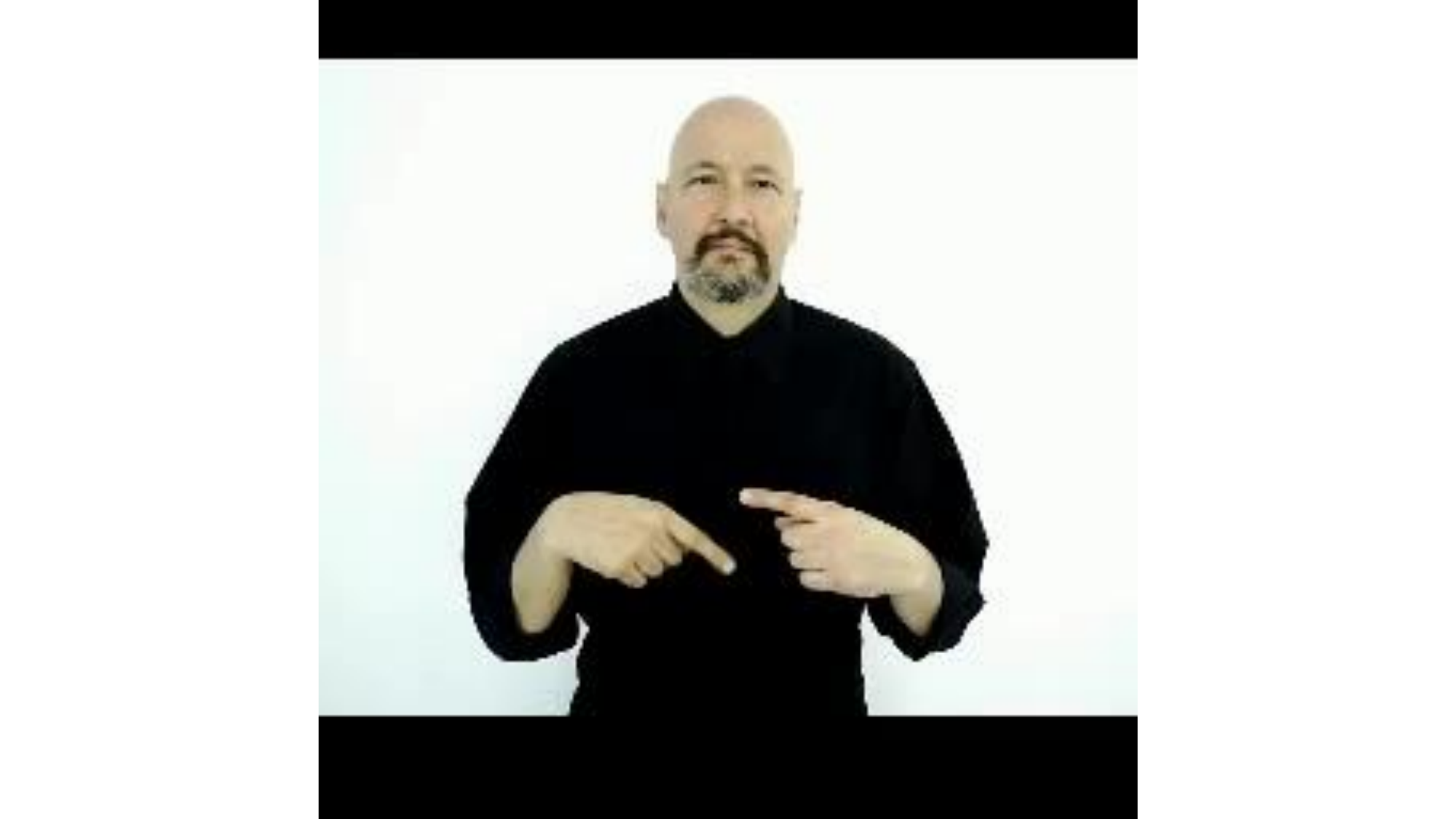}
          \subcaption{WLASL}
          \label{fig:sample_WLASL}
      \end{minipage}

      \begin{minipage}[b]{0.49\hsize}
        \centering
          \includegraphics[width=\hsize]{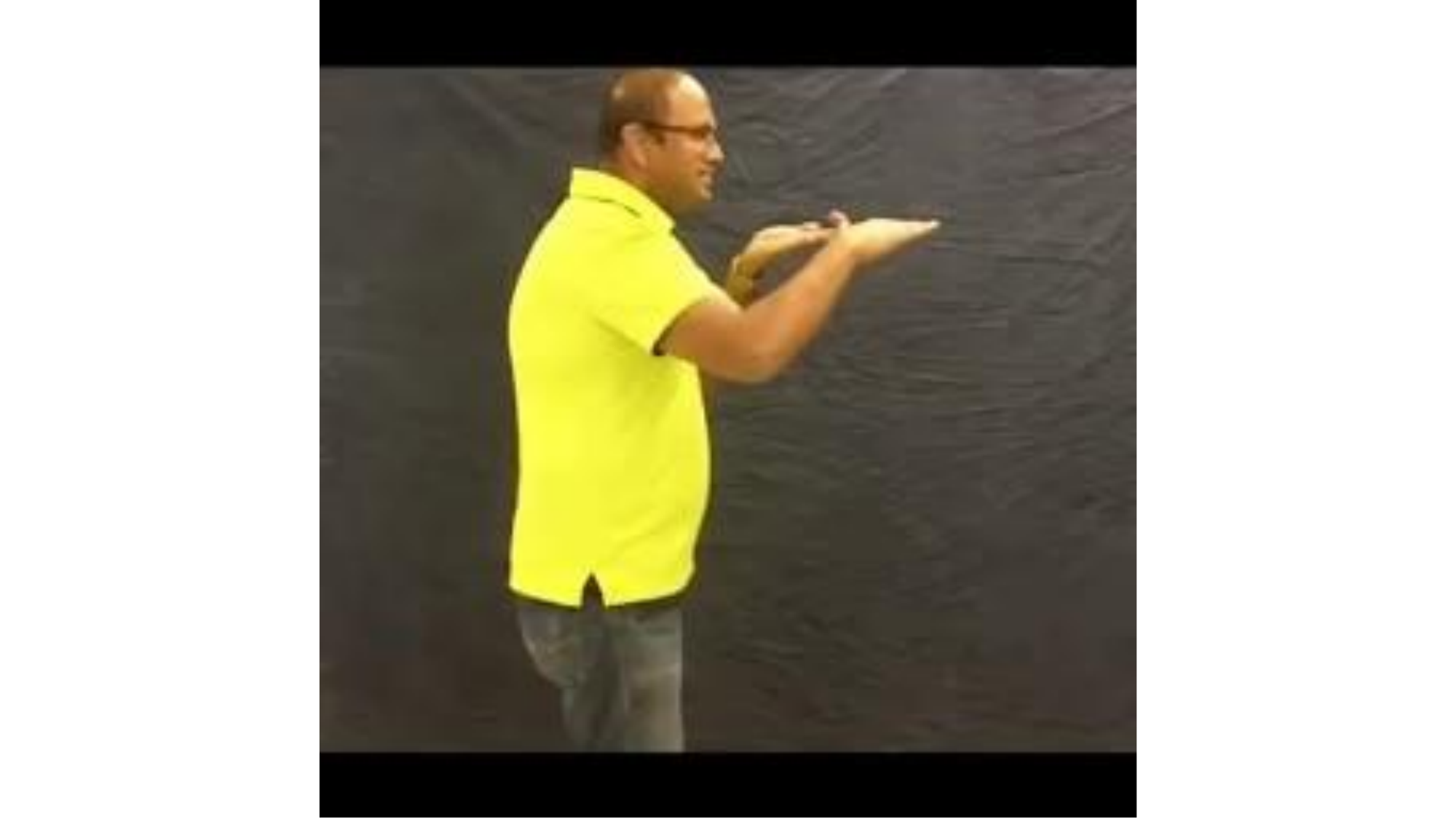}
          \subcaption{MS-ASL (side view)}
          \label{fig:sample_MS-ASL_side}
      \end{minipage}
\hfill
      \begin{minipage}[b]{0.49\hsize}
      \centering
        \includegraphics[width=\hsize]{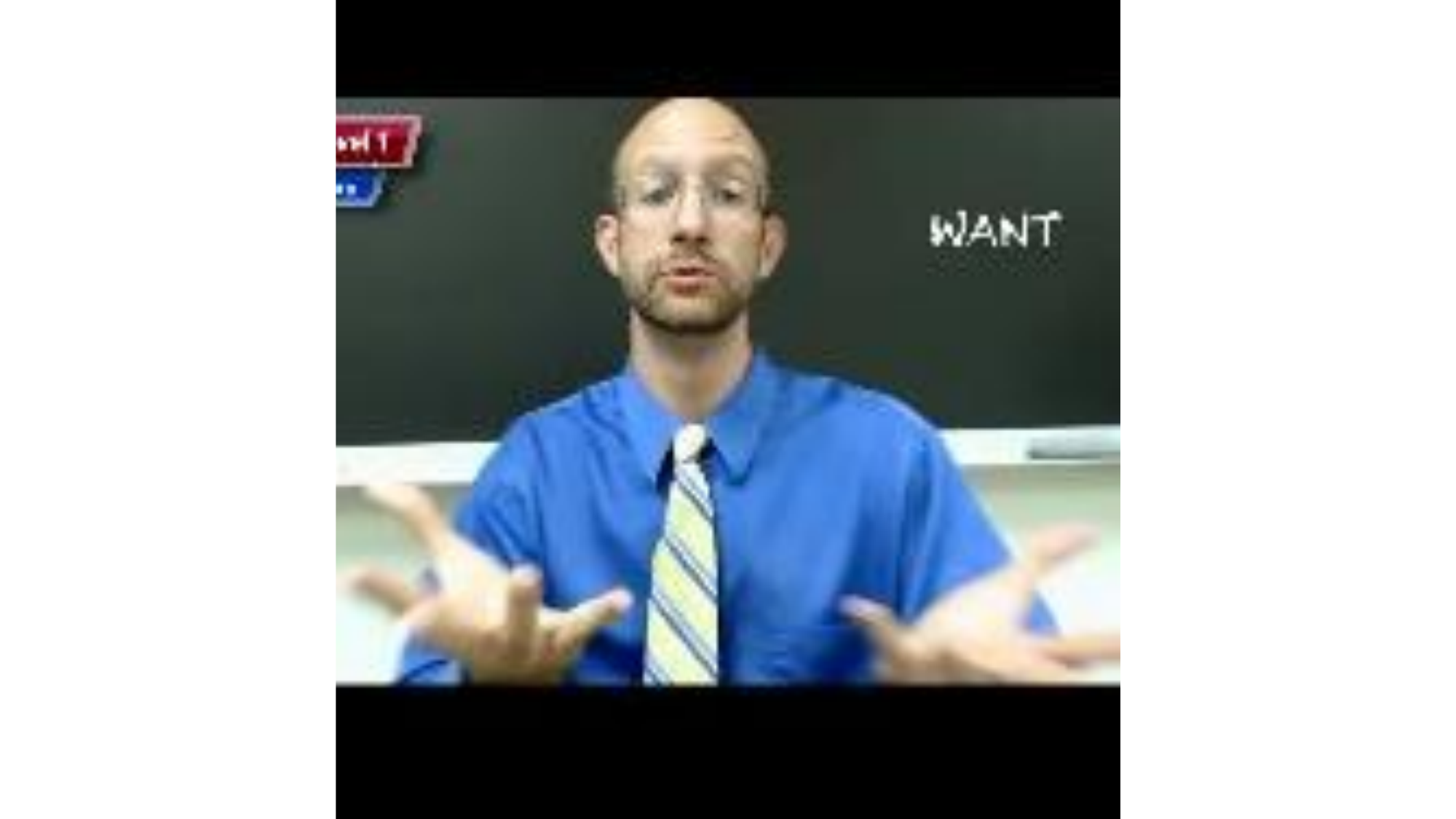}
         \subcaption{MS-ASL (upper body)}
          \label{fig:sample_MS-ASL_upperbody}
      \end{minipage}
      
    \caption{Sample video frames from the WLASL and MS-ASL datasets.}
    \label{fig:sample}
\end{figure}

Similar to the experimental results on the WLASL dataset, we compared the accuracy of the models with and without the local image and skeleton streams.
As a result, we confirmed that the accuracies of the models using both streams were higher than those of not using them as long as they were used alone (i.e., Baseline1 vs Ours1 and Baseline1 vs Ours2).
However, in contrast to the WLASL dataset case, when more than two streams were combined, we could not increase the accuracy in the following cases: Ours5 vs Ours6 on the MS-ASL100 dataset, Baseline2 vs Ours5 on the MS-ASL200 dataset; and Baseline2 vs Ours5, Baseline2 vs Ours6, Ours4 vs Ours5, and Ours4 vs Ours6 on the MS-ASL1000 datasets.
These results confirm that the proposed MSNN was effective for WSLR except in the above cases.
These exceptions seem to be due to the diversity of the data variations contained in the MS-ASL dataset.
\cref{fig:sample} shows some examples of the data in the WLASL and MS-ASL datasets.
As shown in \cref{fig:sample_WLASL}, the WLASL dataset contains video frames whose camera views are limited to a frontal view of the signers and focus on their upper bodies.
In contrast, the MS-ASL dataset contains video frames with different view angles and different body parts, as shown in \cref{fig:sample_MS-ASL_side,fig:sample_MS-ASL_upperbody}.
In the proposed method, three keypoints (i.e., the shoulder, elbow, and wrist) were acquired by OpenPose to detect a hand region.
Therefore, in the case of a close-up of the signer, the elbow point may not be estimated, which causes the extraction of the hand bounding box to fail.
In addition, the cropped local image patches and skeletal estimation results of the videos captured from the side viewpoints are far different from those captured from the front viewpoints.
Hence, we believe that such data affected the training in the local image and skeletal streams and disrupted the SLR accuracy improvement.
In our future work, we therefore need to improve the robustness against view changes, as shown in \cref{fig:sample}, by collecting more training videos captured from various viewpoints.
Moreover, we believe making the extraction of the hand regions robust will be effective, particularly in a close-up view.
The extraction of the hand regions of the signer relies on the elbow point in the current implementation.
However, it would be desirable to extract the hand regions in a way that does not depend on the elbow point.

\begin{figure}[t]
    \centering
    
    \begin{minipage}{\hsize}
        \centering
        \includegraphics[width=\hsize]{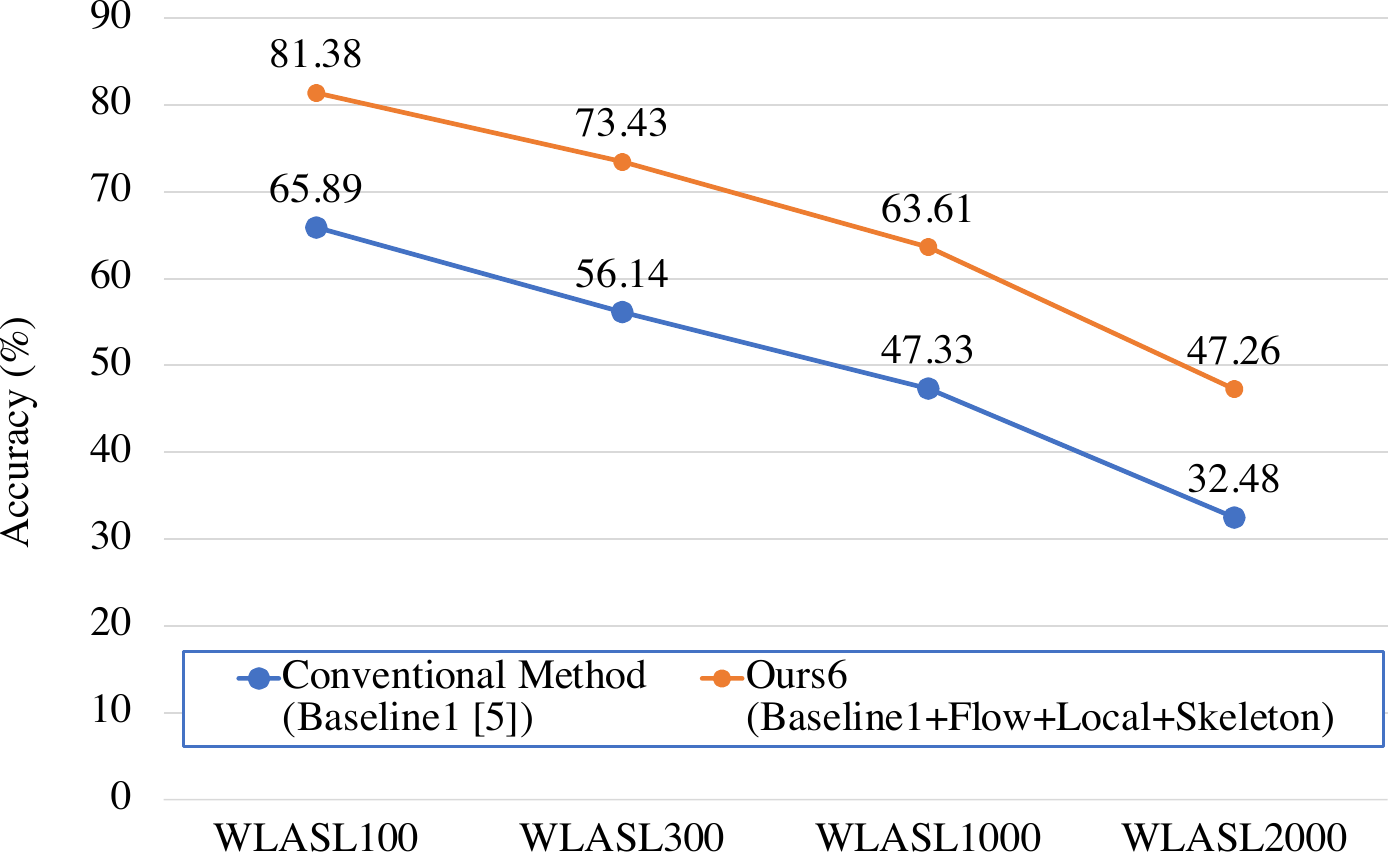}
        \subcaption{WLASL dataset}
        \label{fig:scalability_WLASL}
    \end{minipage}

    
    \begin{minipage}{\hsize}
        \centering
        \includegraphics[width=\hsize]{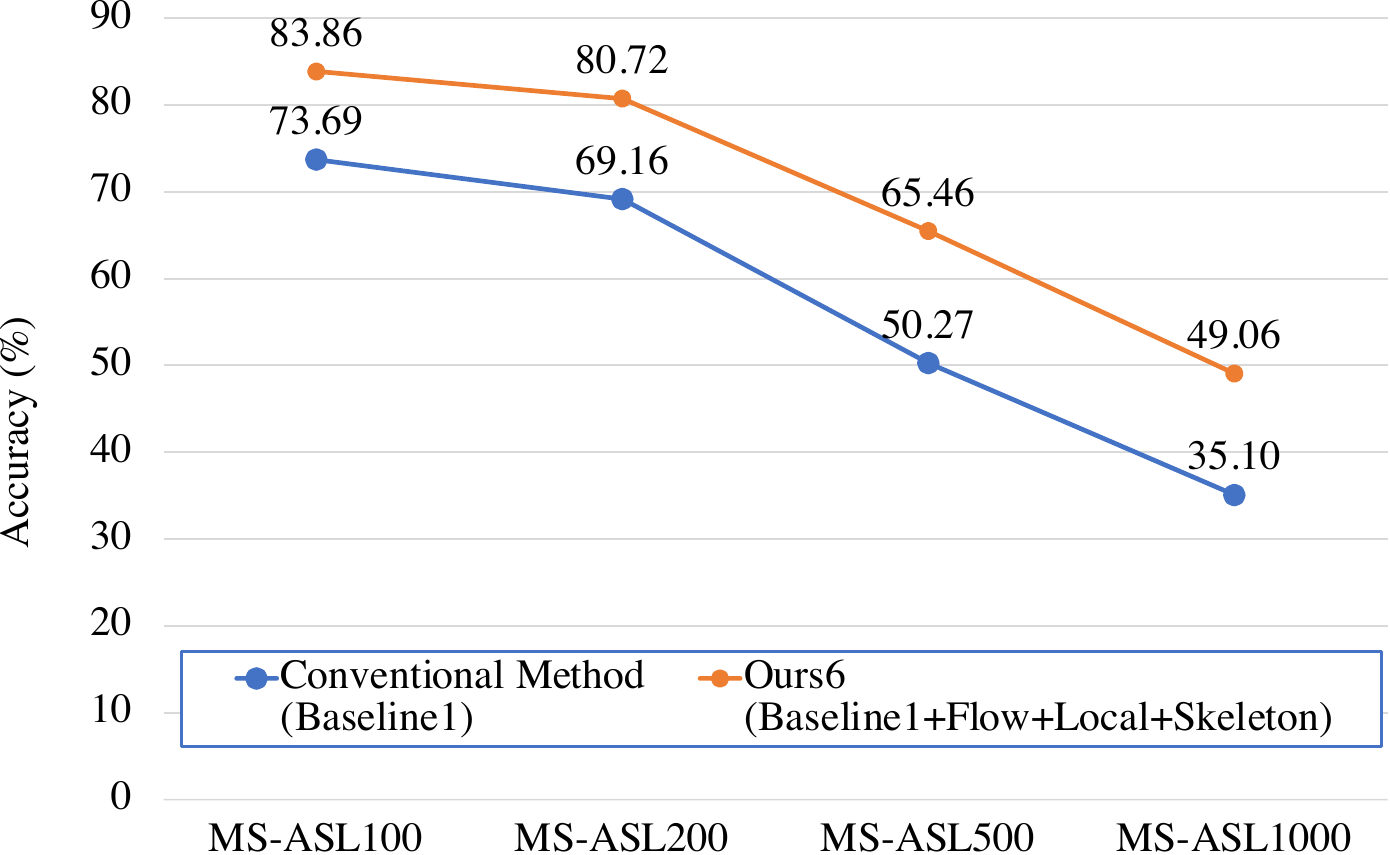}
        \subcaption{MS-ASL dataset}
        \label{fig:scalability_MS-ASL}
    \end{minipage}

    \caption{Scalability of Baseline1 and the proposed method (Ours6) in Top-1 accuracies (\%).}
    \label{fig:scalability}
\end{figure}
\begin{figure}[t]
  \centering
    \includegraphics[width=\hsize]{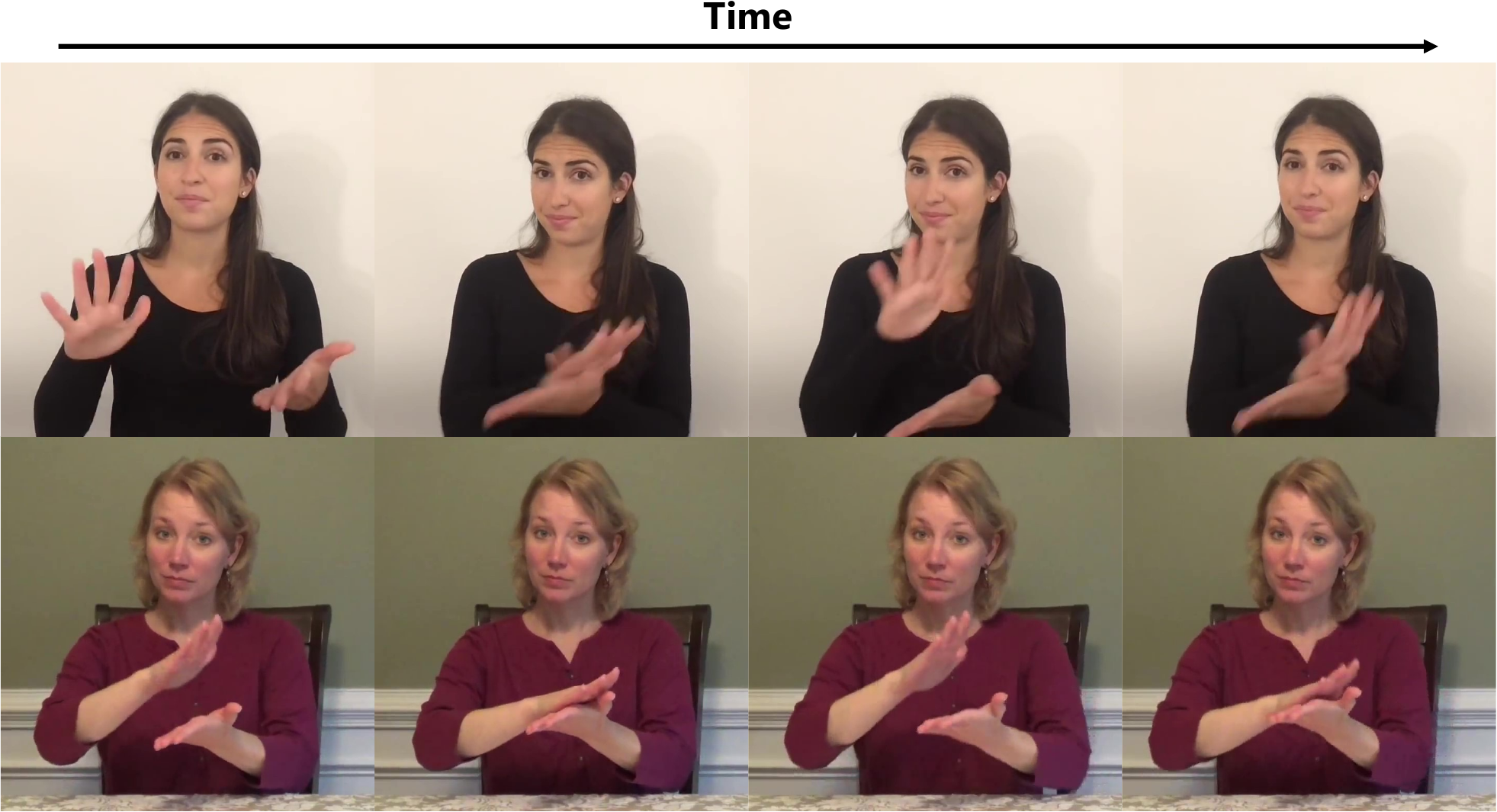}
  \caption{Examples of sign words that are represented by similar gestures:
  (top row) ``paper'' and (bottom row) ``school'' from the MS-ASL dataset~\cite{vaezi2019ms-asl}.}
  \label{fig:ambiguty}
\end{figure}
\vspace{2em}

\begin{figure}[t]
  \centering
    \includegraphics[width=\hsize]{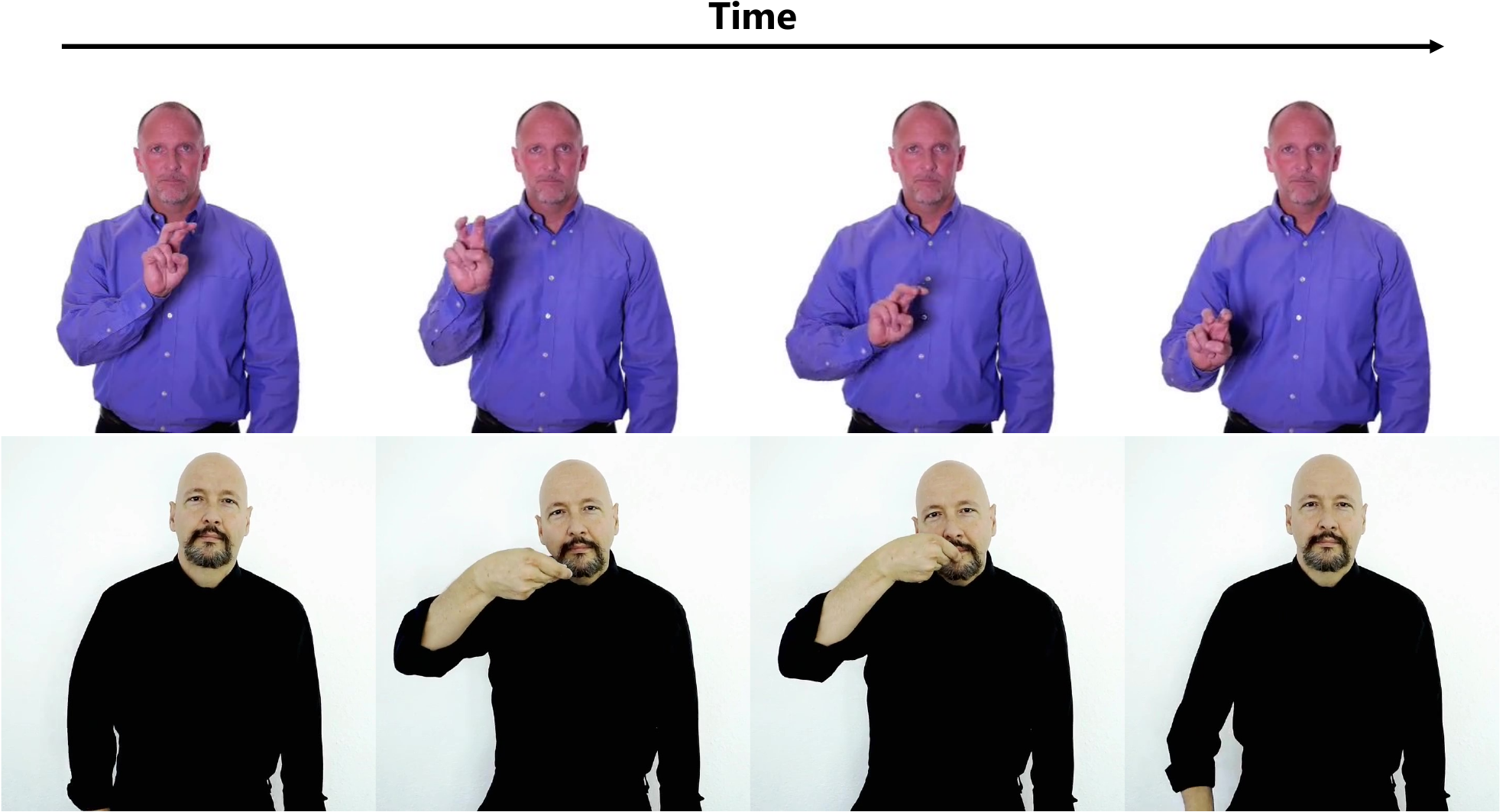}
  \caption{Examples of different gestures that represent the word ``pizza'' from the WLASL dataset~\cite{li2020word}.
  }
  \label{fig:diversity}
\end{figure}

\subsubsection{Scalability on WLASL and MS-ASL datasets}
\cref{fig:scalability} compares the proposed method (Ours6) with Baseline1 on all subsets of the WLASL and MS-ASL datasets.
The graphs show that the proposed method constantly improved the results by approximately 15\% and more than 10\% in Top-1 accuracy compared with respect to the results of Baseline1, regardless of the dataset size on both datasets, respectively.
These results demonstrate the effect of introducing additional information into the WSLR task.
Therefore, it was confirmed that the proposed method exhibited the scalability, 
compared to Baseline1 in terms of superior accuracy regardless of the dataset size.

However, the graphs also show that as the number of classes in the dataset increases, the recognition accuracy of both models decreases monotonically on both datasets.
Top-1 accuracy of the WLASL2000 dataset, which had the closest number of classes to the vocabulary size used in daily life, was as low as 47.26\% even with Ours6, which was the model that achieved the highest accuracy.
Besides, the number of sign words is gradually increasing since new sign words are often created 
if new words are popped up. 
Therefore, from these results, it can be inferred that the proposed method in the current form is not sufficient for practical use in real life.
This result seems to be caused by the relatively small \textit{number of videos per class} of the dataset.
The four subsets of the WLASL dataset consist of Top-100, 300, 1000, and 2000 classes, where the classes were sorted in the ascending order of the number of videos per class.
As shown in \cref{tab:dataset}, the average number of videos per class decreases as the number of classes increases.
In this study, because a supervised learning approach was used for WSLR, more training data can lead to more accurate recognition by enabling richer information to be acquired.
In contrast, similar to the case when a large number of classes in the WLASL and MS-ASL datasets were used, fewer training data may have increased the the following problems.
The first problem is the ambiguity and diversity of expressions in sign language.
As shown in \cref{fig:ambiguty}, there are signs with ambiguous expressions, where similar gestures are used to express different words.
As shown in \cref{fig:diversity}, several different gestures can also represent the same word in sign language.
To solve this problem, we should extract more discriminative information from the images and build a module to combine the information of different gestures  when it is necessary to recognize a word with several gesture representations.
Addressing these problems is our future work.

\subsection{Computational efficiency}\label{computation}
Our primary focus in this study is on the recognition accuracy of isolated SLR with MSNN, and we do not specifically address the computational efficiency of our proposed method. This is consistent with most other papers in the field, except for \cite{Guo_NCA2023}. However, considering the potential applications of our proposed method, we discuss its computational efficiency.
As demonstrated above, Ours6 achieved high accuracy in WSLR by combining multiple DNN models, namely YOLOv3 (or SSD), OpenPose, I3D, and ST-GCN. This results in Ours6 becoming a large network, which can be a drawback. However, a large network does not necessarily hinder the possibility of real-time applications. Some DNN models used in the proposed method (i.e., YOLOv3, SSD, and OpenPose) can operate in near real-time when run individually and can be executed in parallel. This suggests that even though Ours6 comprises multiple DNN models, it does not necessarily consume an excessive amount of time during inference.
It is important to note that isolated SLR system applications are not limited to real-time scenarios. There are situations where real-time processing is not required, such as offline batch processing. Hence, even if Ours6 does not run in real-time, we believe our method offers significant value.

Similar to the above the inference speed problem, the computational resources are also important factor. 
Since the models employed in our method require a quite high-spec computational machines, 
this requirement is a hurdle to realize stress-free communication 
between the people with speech impairment and those who can hear. 
This is also one of the remaining limitations of our method. 
To solve this problem, we need to create lightweight DNN models for WSLR 
by utilizing model compression techniques such as model pruning and knowledge distillation, 
while keeping the recognition accuracy. 
This is a part of our future work.


\section{Conclusion}\label{conclusion}
In this paper, a method with a multi-stream structure focusing on global information, local information, and skeletal information to improve the accuracy of WSLR was proposed. The proposed multi-stream structure consists of base, local image, and skeleton streams. The base stream is used in conventional methods. However, it does not exploit the local information of the hands and face nor the relative positions of the body and both hands. Hence, two other streams were introduced. The local image stream captures local information such as handshape and facial expression. The skeleton stream captures hand position relative to the body. 
By combining these three streams, the proposed method achieves higher recognition accuracy than the state-of-the-art methods. In particular, the experimental results on the WLASL dataset confirmed that the proposed multi-stream models achieved 81.38\%, 73.43\%, 63.61\%, and 47.26\% in Top-1 accuracy on the WLASL100, WLASL300, WLASL1000, and WLASL2000 datasets, respectively, which are higher than the results of the conventional methods using only global information. Moreover, to verify the effectiveness of the local image and skeleton streams, the recognition accuracy of the model with and without each stream was compared. As a result, the models with all three streams achieved higher recognition accuracies than the other models at the time our pre-print was released. This confirms that the three streams used in the proposed method were effective for WSLR. Moreover, in the experiments on the MS-ASL dataset, the proposed method achieved 83.86\%, 80.72\%, 65.46\%, and 49.06\% in Top-1 accuracy on the MS-ASL100, MS-ASL200, MS-ASL500, and MS-ASL1000 datasets, respectively. These results were better than those of the conventional methods on all datasets except for the MS-ASL100 and MS-ASL1000 datasets. Therefore, these results confirm that the proposed method is not a data-specific method, but a highly versatile method for WSLR.

In the future, our aim is to further improve the recognition accuracy by designing a model that can handle various types of information. 
Besides, in this study, we investigated the effectiveness of our proposed method on the WLASL and MSASL datasets. 
The videos in both datasets were captured in relatively controlled environments. 
Therefore, future work includes further investigation into more realistic environments 
that may contain various signers and complex and noisy backgrounds.
In addition, the proposed model should be able to be applied to sign languages other than ASL such as British, Japanese, and Indian sign languages. Hence, we plan to perform additional experiments using datasets of these sign languages 
and explore the generalizability of the proposed method to other sign languages 
along with any necessary modifications based on each sign language. 
This is also a part of our future work. 
Besides, since the number of sign language words is gradually increasing, 
we need to improve the scalability capability of our method. 
This is one of our future work. 
Additionally, with an eye towards real-time applications, addressing the time consumption and the required resources of our method is a part of our future work.
After solving these work, it is desired that our framework is extended to handle continuous sign language recognition for support hearing-impaired people in the future.

\EOD

\end{document}